\documentclass{article}

\usepackage[dblblindworkshop, final]{neurips_2025}

\usepackage[utf8]{inputenc} 
\usepackage[T1]{fontenc}    
\usepackage{hyperref}       
\usepackage{url}            
\usepackage{booktabs}       
\usepackage{amsfonts}       
\usepackage{nicefrac}       
\usepackage{microtype}      
\usepackage{xcolor}         
\usepackage{graphicx}
\usepackage{color}

\usepackage{amssymb}
\usepackage{float}
\usepackage{multirow}
\usepackage{makecell}
\usepackage{lipsum}
\usepackage{bm}
\usepackage{subcaption}
\usepackage{hyperref}
\usepackage{tcolorbox}
\tcbuselibrary{breakable, skins}

\workshoptitle{Efficient Reasoning}

\title{Finding the Sweet Spot: Trading Quality, Cost, and Speed During Inference-Time LLM Reflection}

\author{%
  Jack Butler \\
  Amazon Web Services \\
  \texttt{jackbtlr@amazon.co.uk} \\
  \And
  Nikita Kozodoi \\
  Amazon Web Services \\
  \texttt{kozodoi@amazon.de} \\
  \And
  Zainab Afolabi \\
  Amazon Web Services \\
  \texttt{zafolabi@amazon.co.uk} \\
  \And
  Brian Tyacke \\
  Zalando \\
  \texttt{brian.tyacke@zalando.de} \\
  \And
  Gaiar Baimuratov \\
  Zalando \\
  \texttt{gaiar.baimuratov@zalando.de} \\
}

\begin{document}

\maketitle

\begin{abstract}
As Large Language Models (LLMs) continue to evolve, practitioners face increasing options for enhancing inference-time performance without model retraining, including budget tuning and multi-step techniques like self-reflection. While these methods improve output quality, they create complex trade-offs among accuracy, cost, and latency that remain poorly understood across different domains. This paper systematically compares self-reflection and budget tuning across mathematical reasoning and translation tasks. We evaluate prominent LLMs, including Anthropic Claude, Amazon Nova, and Mistral families, along with other models under varying reflection depths and compute budgets to derive Pareto optimal performance frontiers. Our analysis reveals substantial domain dependent variation in self-reflection effectiveness, with performance gains up to 220\% in mathematical reasoning. We further investigate how reflection round depth and feedback mechanism quality influence performance across model families. To validate our findings in a real-world setting, we deploy a self-reflection enhanced marketing content localisation system at Lounge by Zalando, where it shows market-dependent effectiveness, reinforcing the importance of domain specific evaluation when deploying these techniques. Our results provide actionable guidance for selecting optimal inference strategies given specific domains and resource constraints. We open source our self-reflection implementation for reproducibility at \url{https://github.com/aws-samples/sample-genai-reflection-for-bedrock}.
\end{abstract}


%
%

%
%

\section{Introduction}
\label{sec_introduction}

The deployment of large language models (LLMs) in production systems has created new challenges for practitioners seeking to optimise performance under real-world constraints. Recent advances in inference-time
computation scaling \cite{snell2025scaling} offer promising solutions, allowing deployed systems to dynamically allocate computational resources based on task difficulty, available budget, and latency requirements
without model retraining.

Two established approaches for adjusting inference-time resources are multi-step inference and budget tuning. Budget tuning, available for select LLMs such as Anthropic Claude 3.7 Sonnet and OpenAI o1, enables users to configure inference parameters like maximum thinking tokens or reasoning tiers (e.g., low or high), allocating greater computational effort to more challenging inputs. Multi-step approaches such as self-reflection \cite{madaan2023selfrefine} are model-agnostic and involve prompting a model to revise its responses through sequential follow-up calls to the model.

Prior work demonstrates that self-reflection improves LLM performance in tasks with clearly defined evaluation criteria \cite{madaan2023selfrefine} and structured domains with informative feedback signals, such as programming or math \cite{chen2024teaching}. For instance, executing generated code and providing the outputs back to the LLM as context creates concrete feedback signals for accurate self-reflection. 
However, most production applications such as translation or classification involve more ambiguous objectives and weaker feedback signals. The effectiveness of self-reflection on such tasks remains underexplored, making it unclear whether additional inference-time computation consistently yields performance gains across diverse domains.

This uncertainty is compounded by the rapidly evolving LLM landscape, where practitioners navigate dozens of model options across providers, each with different capabilities, pricing structures, and
inference features. Production teams frequently face critical decisions: Should they deploy a smaller, cost-effective model with advanced inference strategies, or invest in larger models with simpler inference
pipelines? How do these choices impact not just accuracy, but operational costs, latency requirements, and system reliability?

This paper addresses these practical deployment challenges through comprehensive evaluation of inference-time optimisation strategies across real-world applications. We make three primary contributions. First, we benchmark self-reflection and budget tuning across multiple LLMs and application domains including mathematical reasoning, text-to-SQL generation, sentiment classification, and translation. The derived Pareto-optimal frontiers illustrate accuracy-latency trade-offs of different strategies and provide actionable recommendations for selecting a suitable inference method based on domain-specific requirements, resource constraints, and the base model. Second, we analyse self-reflection trajectories across different LLMs and feedback mechanisms, revealing how reflection depth and feedback quality critically influence self-reflection performance. To the best of our knowledge, this work presents the first direct performance comparison between these two approaches. Third, we demonstrate our findings through a production deployment at Lounge by Zalando, where we implemented self-reflection for marketing content localisation across 17 European markets. 
\section{Related Work}
\label{sec_related_work}

\subsection{Inference-Time Compute}

LLMs such as Anthropic's Claude family \cite{claude3} have been increasing in size over recent years, which has brought profound improvements in performance across a wide range of applications while simultaneously increasing training time and compute requirements \cite{hoffmann2022trainingcomputeoptimallargelanguage}. Inference-time compute optimisation techniques avoid modifying the pre-trained model and instead enable dynamic allocation of computational resources depending on the specific requirements of each input. This offers the ability to tune performance according to task demands, scaling performance at inference time \cite{snell2025scaling}.

One of the prominent inference optimisation approaches is self-reflection \cite{madaan2023selfrefine}, which performs sequential follow-up LLM calls, allowing it to revise initial responses. Other studies have explored drawing parallel samples from the LLM and implementing sampling and verification procedures such as tree-of-thoughts \cite{yao2023treethoughtsdeliberateproblem} and graph-of-thoughts \cite{Besta_2024}. Recent work has also leveraged techniques such as temporary fine-tuning \cite{akyürek2025surprisingeffectivenesstesttimetraining} and nearest neighbour retrieval-based fine-tuning \cite{hubotter2025efficiently} where model's parameters are temporarily updated during inference.

In this paper, we explore trade-offs between two established inference strategies: model-agnostic self-reflection \cite{madaan2023selfrefine} and built-in reasoning capabilities exposed through model provider APIs. Our goal is to provide insights for practitioners who lack resources to conduct large-scale per-sample fine-tuning or complex multi-step inference processes with feedback loops.

\subsection{LLM Post-Training}

Another research area aimed at LLM reasoning capabilities is the use of reinforcement learning on specialised reasoning datasets. These approaches implement reinforcement learning with access to either outcome supervision \cite{trung-etal-2024-reft,r3}, step-by-step process supervision \cite{lightman2023letsverifystepstep,setlur2025rewarding} or LLM-driven feedback mechanisms \cite{lee2024rlaif}. 

Reasoning models are commonly deployed via API interfaces with configuration settings that allow users to adjust computational resources allocated per sample (e.g. Anthropic Claude 3.7 Sonnet thinking tokens budget). Crucially, the reasoning in these models happens as internal processing tokens, and discrete proposed solutions are not validated using external feedback during generation.

\subsection{LLM Evaluation}

As LLMs have become more capable and businesses increasingly incorporate them into their products and services, robust evaluation frameworks have become essential. Various evaluation platforms exist across different domains, such as HELM \cite{liang2023holisticevaluationlanguagemodels} and Chatbot Arena \cite{chiang2024chatbotarenaopenplatform}, where models undergo evaluation and receive automated or human feedback scores depending on the task and domain. 

While these platforms provide standardised evaluation of different LLMs, including base models and fine-tuned or quantised variants, there is a lack of comparable evaluation frameworks for inference techniques. This gap makes it challenging for practitioners to understand and navigate trade-offs when combining LLMs with various inference-time compute methods. Throughout our experimentation, we demonstrate these trade-offs across model families, task domains, and inference budgets.
%
%

\section{Experimental Setup}
\label{sec_architecture}


\subsection{Datasets}

The empirical study splits into two stages: testing on established benchmarks followed by evaluation on real-world deployment application. First, we perform experiments across four distinct domains using established benchmarks:

\begin{itemize}
    \item \textbf{Flores-200 (Translation)} \cite{nllb2022}: Multilingual translation benchmark spanning 200 languages, allowing assessment of cross-lingual capabilities.
    
    \item \textbf{Math500 (Mathematical reasoning)} \cite{lightman2024lets}: Dataset with 500 problems across algebra, arithmetic and more, testing symbolic manipulation and logical reasoning.
    
    \item \textbf{Spider (Text-to-SQL)} \cite{yu-etal-2018-spider}: Complex text-to-SQL task involving 200 databases with multiple tables, evaluating structured SQL query generation capabilities.
    
    \item \textbf{IMDB Reviews (Sentiment analysis)} \cite{maas-EtAl:2011:ACL-HLT2011}: Binary sentiment classification dataset on movie reviews, assessing natural language classification performance.
\end{itemize}

The diverse set of tasks allows us to evaluate structured mathematical and programming reasoning (Math500, Spider) and natural language understanding (Flores-200, IMDB). We use a subset of each dataset for evaluation. For Spider, we use 5 databases (voter\_1, battle\_death, museum\_visits, employee\_hire\_evaluation, orchestra). For Flores-200, we sample 200 examples across 15 language pairs to ensure cross-linguistic variation (English to Arabic, German, Spanish, French, Hebrew, Hindi, Italian, Japanese, Korean, Dutch, Portuguese, Russian, Turkish, Chinese, Polish). For IMDB and Math500, we randomly sample 100 examples.

After validating approaches on benchmark data sets, we test and deploy the best model configurations on real-world data on marketing content localisation at Lounge by Zalando. Further details on the deployment are provided in Section \ref{sec:case_study}.

\subsection{Models and Inference Techniques}

We compare performance across 10 LLMs: 

\begin{itemize}
    \item Amazon Nova (Premier, Pro, Micro, Lite \cite{Intelligence2024}) \item Anthropic Claude (Sonnet 3.7, Sonnet 3.5 v2, Haiku 3.5 \cite{claude3})
    \item Llama 4 (Maverick 17B \cite{Llama4}) \item Mistral (Small, Large \cite{Mistral}).
\end{itemize}

For inference, we employ self-reflection with 0, 1, and 3 reflection rounds across all LLMs. Self-reflection is implemented through repeated LLM invocations, where at the end of each round we prompt the model to reflect on its response and update it if necessary.
On the text-to-SQL task, we also compare 2 feedback mechanisms, which provide additional context before each self-reflection round. For Claude 3.7, we additionally make use of the built-in reasoning mode by defining 2 thinking budgets (4096 tokens and 1024 tokens). We refer to these thinking budgets as high and low in the rest of the paper.

We run experiments using Amazon Bedrock, maintaining default temperature and inference parameters associated with each model to ensure fair comparison. The token costs are recorded as of 02/05/2025, assuming on-demand pricing. Latency is measured as the total elapsed time between the input and the completion of the full response. Prompts used for each of the 4 domains, as well as for self-reflection and feedback mechanisms, are available in Appendix A.

\subsection{Evaluation Metrics}

We employ task-specific metrics for each dataset to evaluate LLM performance. For translation, we use METEOR \cite{meteor}, which accounts for both precision and recall while handling synonyms and paraphrases. Sentiment analysis quality is assessed using classification accuracy on the binary prediction task. Spider and Math500 use additional verification procedures.

For Spider and Math500, we implement additional verification procedures to assess semantic equivalence of generated responses. 
For Math500, we evaluate accuracy through normalised comparison and symbolic verification. We use string matching on normalised and cleaned LaTeX expressions, followed by symbolic equivalence checking with SymPy \cite{sympy} to identify equivalent answers even when expressed differently. For Spider, we assess both exact matches and functional equivalence by executing SQL queries, discarding failures, and comparing normalised result tables against ground truth. When exact row matches are not found, we calculate partial credit based on matching cell values.

For the content localisation deployment data, we use multiple technical metrics to evaluate the localisation quality, including METEOR, BLEU, and LLM-as-a-judge score. After technical evaluation, we also run human expert evaluation tests of the deployed system, where we compare the number of localisation mistakes raised by expert copywriters after analysing the generated localisations.
\section{Results on Benchmarks}
\label{sec_inputs}

This section overviews empirical results. For each dataset, we illustrate the percentage change in accuracy relative to zero reflections for each model, highlighting improvements for each configuration. Next, we construct Pareto-optimal frontiers showing accuracy-latency trade-off when employing inference strategies and provide cost information for each model and strategy combination. In Section \ref{sec:ablations}, we dive deeper on how performance of self-reflection is affected by different factors.

\subsection{Mathematical Reasoning (Math500)}

\begin{figure}[H]
\begin{subfigure}[b]{0.45\textwidth}
    \includegraphics[width=\textwidth,trim={0 0.25cm 0 0},clip]{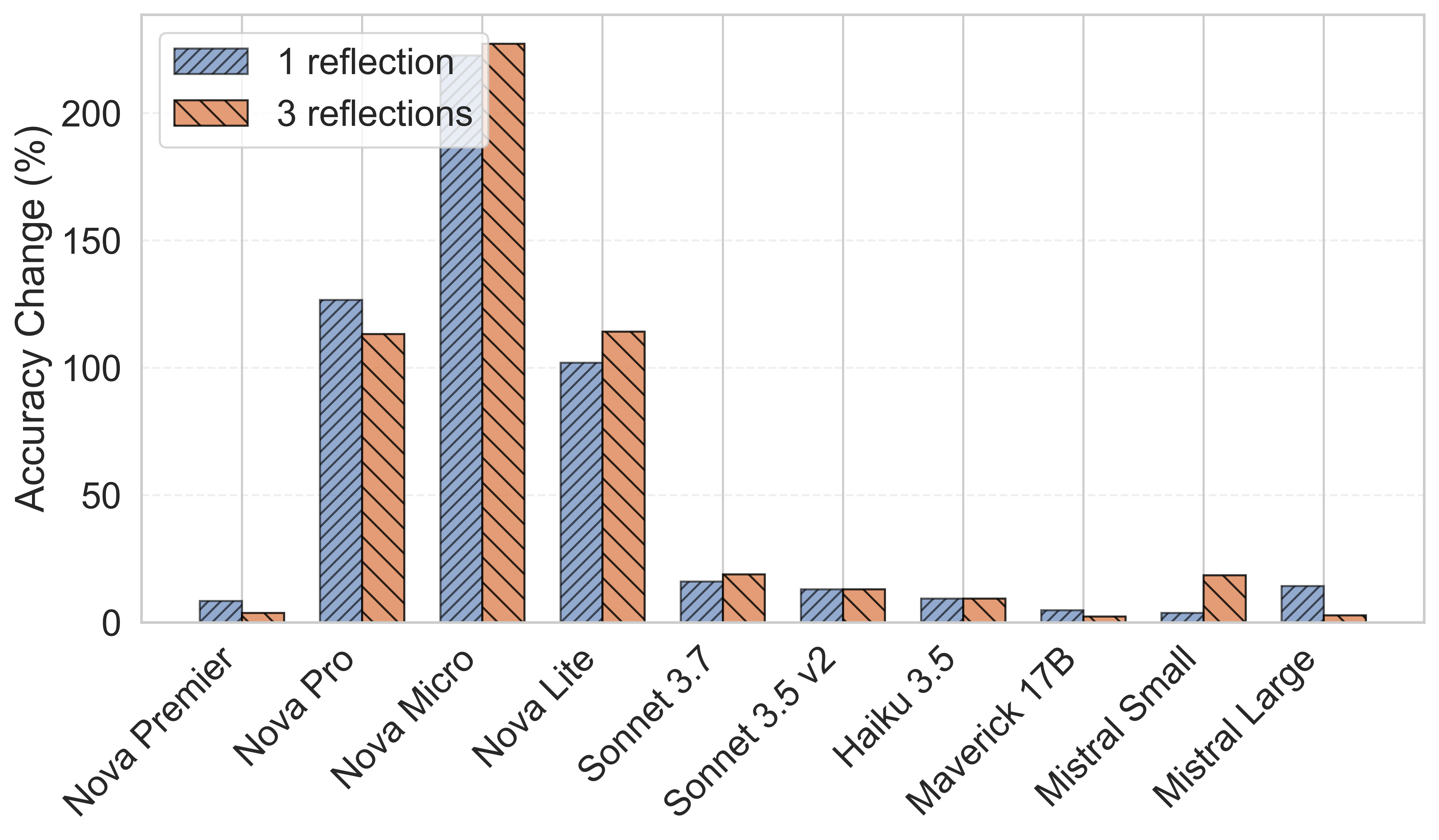}
    \caption{Relative Self-Reflection Gains}
\end{subfigure}
\hfill
\begin{subfigure}[b]{0.45\textwidth}
    \includegraphics[width=\textwidth,trim={0 0.25cm 0 0},clip]{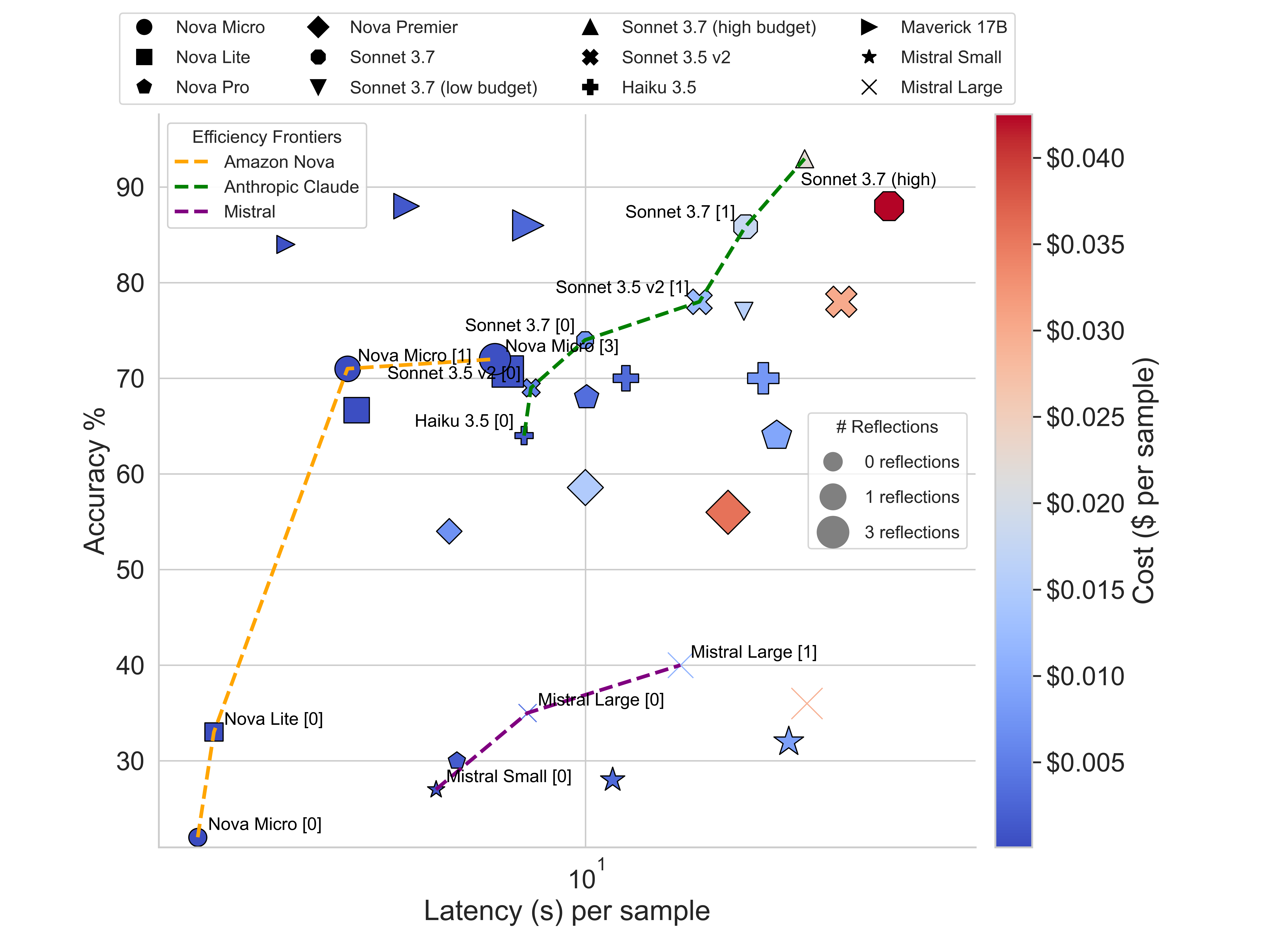}
    \caption{Accuracy-Latency Pareto Frontiers}
\end{subfigure}
\caption{Inference-Time Performance (Math500)}
\label{fig:math500}
\end{figure}

As depicted on Figure \ref{fig:math500}(a), all LLMs benefit from self-reflection. Nova Micro shows the largest gains, with accuracy improving by 220\% with 1 reflection and maintaining this gain after 3 reflections. This suggests that Nova Micro's base mathematical reasoning capabilities are significantly enhanced through iterative self-correction. Similarly, Nova Lite and Pro show substantial improvements of approximately 100-130\% with reflection, indicating that smaller models in the Nova family particularly benefit from reflection in mathematical reasoning.

In contrast, Nova Premier and Claude LLMs exhibit more modest but consistent improvements from reflection. Sonnet 3.7 shows approximately a 16\% increase in accuracy with 1 and 20\% with 3 reflections. Sonnet 3.5 v2 and Haiku 3.5 demonstrate similar patterns with gains of 13\% and 9\%, respectively. These results suggest that while Claude models benefit from reflection in mathematical tasks, their initial performance is already strong, resulting in less dramatic relative improvements.

Figure \ref{fig:math500}(b) provides accuracy measurements across model configurations. Overall, Sonnet 3.7 demonstrates superior mathematical reasoning capabilities, starting with a baseline accuracy of 74\% without reflection and improving to 86\% with 1 and 88\% with 3 reflections. Nova Micro starts with at just 22\% accuracy without reflection (omitted from the plot) but jumps to 71\% accuracy with 1 and 72\% with 3 reflections. This pattern is similar for other Nova, Llama Maverick, Mistral and Claude LLMs and suggests that for mathematical reasoning, a single well-implemented reflection round captures most of the potential performance benefit, with diminishing returns for further rounds.

The Pareto frontier for the Claude family offers a rich selection in the accuracy-latency space, ranging from Haiku 3.5 with no reflection offering 64\% accuracy at \$0.0015 per example and 7.5 sec. latency, up to Sonnet 3.7 with a high thinking budget, which reaches 93\% accuracy at \$0.0224 and 27.9 sec. latency. At the same time, Sonnet 3.7 with a low thinking budget is dominated by Sonnet 3.7 with 1 self-reflection, which reaches a higher accuracy at the same latency. Considering the Amazon Nova family, Nova Micro with 1 and 3 reflections dominate Haiku 3.5 and Sonnet 3.5 in low-latency space but can not reach the same accuracy as higher-end Claude models even with self-reflection. This implies that practitioners should consider Nova Micro with self-reflection under strict cost/latency constraints and switch to Sonnet 3.7 with high reasoning for the best performance. Llama Maverick provides superior accuracy across the Nova Frontier for budgets larger than Nova Lite or Micro without reflections, and also match the performance of Sonnet 3.7 with 1 reflection.

To validate statistical significance of the results, we draw 100 bootstrap samples of individual examples and run pairwise t-tests comparing mean accuracies per model configuration. All models on the efficiency frontiers show significant accuracy differences at 1\% level, which are also confirmed by pairwise Nemenyi tests. 

\subsection{Text-to-SQL (Spider)}

In contrast to Math500, in text-to-SQL generation Sonnet 3.7 is the only model to show consistent and limited improvements with additional reflections, gaining 2.3\% accuracy with 1 round and a 5.6\% gain with 3 rounds. Most other LLMs show mixed or negative responses to self-reflection. Sonnet 3.5 v2 demonstrates the most pronounced quality degradation, with accuracy declining by approximately 4.8\% with 1 and reflection rounds. Similarly, Nova Pro and Haiku 3.5 show noticeable performance decreases with added reflection rounds.

Nova Micro, Sonnet 3.7 and Llama Maverick are the only models benefiting from additional reflections. Nova Micro maintains neutral performance with 1 round and achieves a 2.2\% accuracy improvement with 3 reflections. Nova Lite displays an inconsistent pattern, with a slight improvement (1.5\%) at 1 reflection but declining by 1.5\% with 3 rounds. Overall, these results suggest that self-reflection is less useful in domains like text-to-SQL, where revising the generated query without any additional context may mislead LLMs to change previously correct SQL queries. Mistral Small shows benefits with three rounds of reflection but decreases in accuracy with 1 reflection whereas the Large variant demonstrates the opposite behaviour.

\begin{figure}[H]
\begin{subfigure}[b]{0.45\textwidth}
    \includegraphics[width=\textwidth,trim={0 0.25cm 0 0},clip]{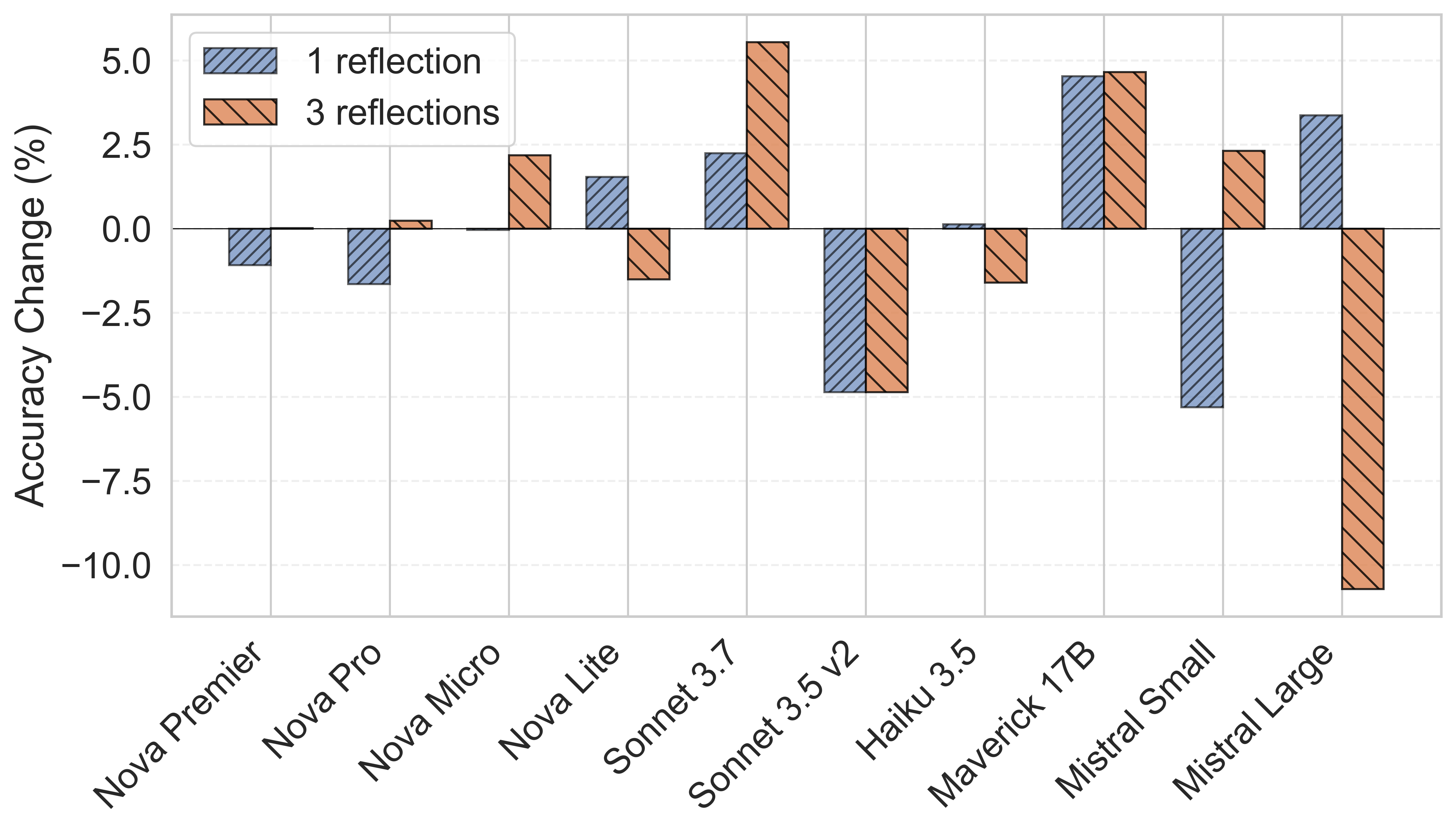}
    \caption{Relative Self-Reflection Gains}
\end{subfigure}
\hfill
\begin{subfigure}[b]{0.45\textwidth}
    \includegraphics[width=\textwidth,trim={0 0.25cm 0 1.5cm},clip]{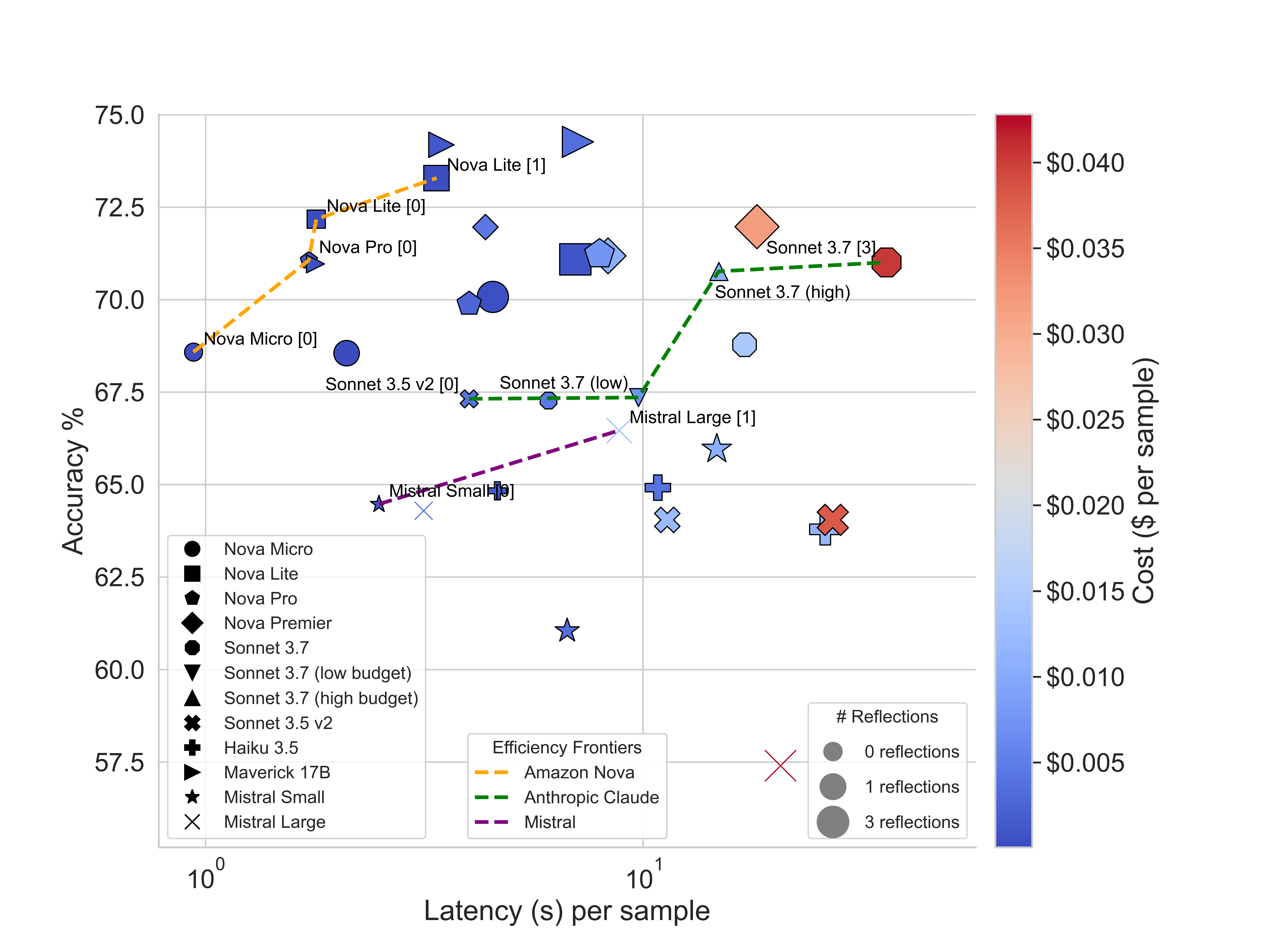}
    \caption{Accuracy-Latency Pareto Frontiers}
\end{subfigure}
\caption{Inference-Time Performance (Spider)}
\label{fig:spider}
\end{figure}
\vspace*{-1ex}

The Amazon Nova Pareto frontier on Figure \ref{fig:spider}(b)  represents optimal configurations for lower-latency applications. Overall, Amazon Nova models consistently outperform Claude LLMs, with 2  Nova Lite variants dominating all Claude counterparts. Nova Lite with 1 reflection achieves the highest absolute accuracy (74\%) with approximately 3-second latency, while Nova Micro with 0 reflections offers the fastest and cheapest option to reach 68\% accuracy. The Claude frontier represents a different set of trade-offs, with Sonnet 3.7 using 3 reflections achieving 71\% accuracy but at a substantially higher latency (>35 seconds) and cost. Interestingly, built-in reasoning modes with both budget sizes fall behind the model variant with 3 reflections in terms of the accuracy, but are available at a lower latency and cost. Mistral frontier highlights that Mistral Small is performant for this use case but to achieve better performance one can leverage Mistral Large with 1 reflection. Llama Maverick provides the highest accuracy and also deliver this at a equivalent latency and cost the Nova Lite.

These results highlight the importance of model-specific optimisation strategies for SQL tasks, with Amazon Nova models generally performing best with minimal reflection, while Claude Sonnet 3.7 uniquely benefits from both reflection and built-in reasoning despite the increased latency and cost.

\subsection{Sentiment Classification (IMDB)}

\begin{figure}[ht]
\begin{subfigure}[b]{0.45\textwidth}
    \includegraphics[width=\textwidth,trim={0 0.25cm 0 0},clip]{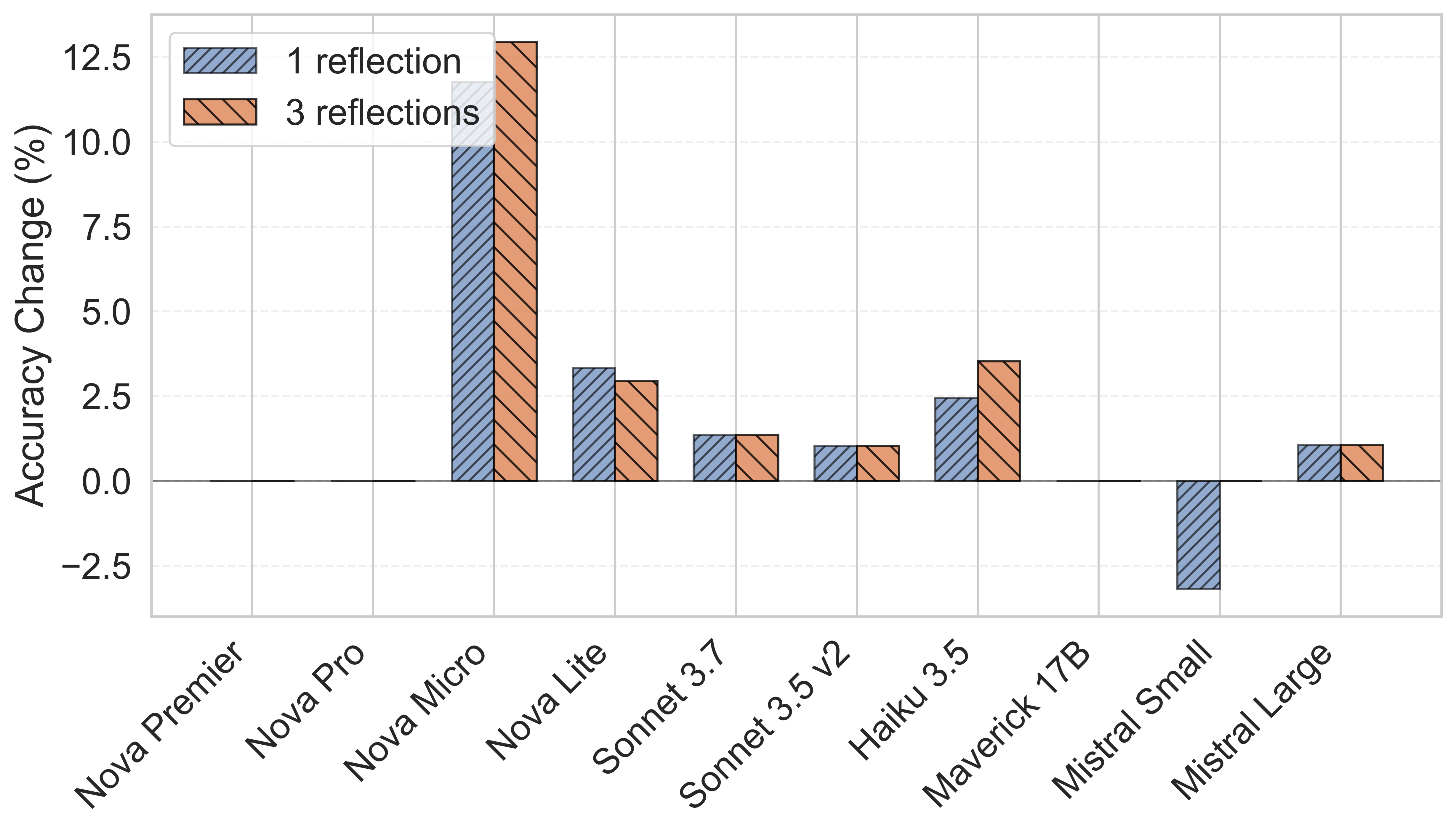}
    \caption{Relative Self-Reflection Gains}
\end{subfigure}
\hfill
\begin{subfigure}[b]{0.45\textwidth}
    \centering
    \includegraphics[width=\textwidth,trim={0 0.25cm 0 1.5cm},clip]{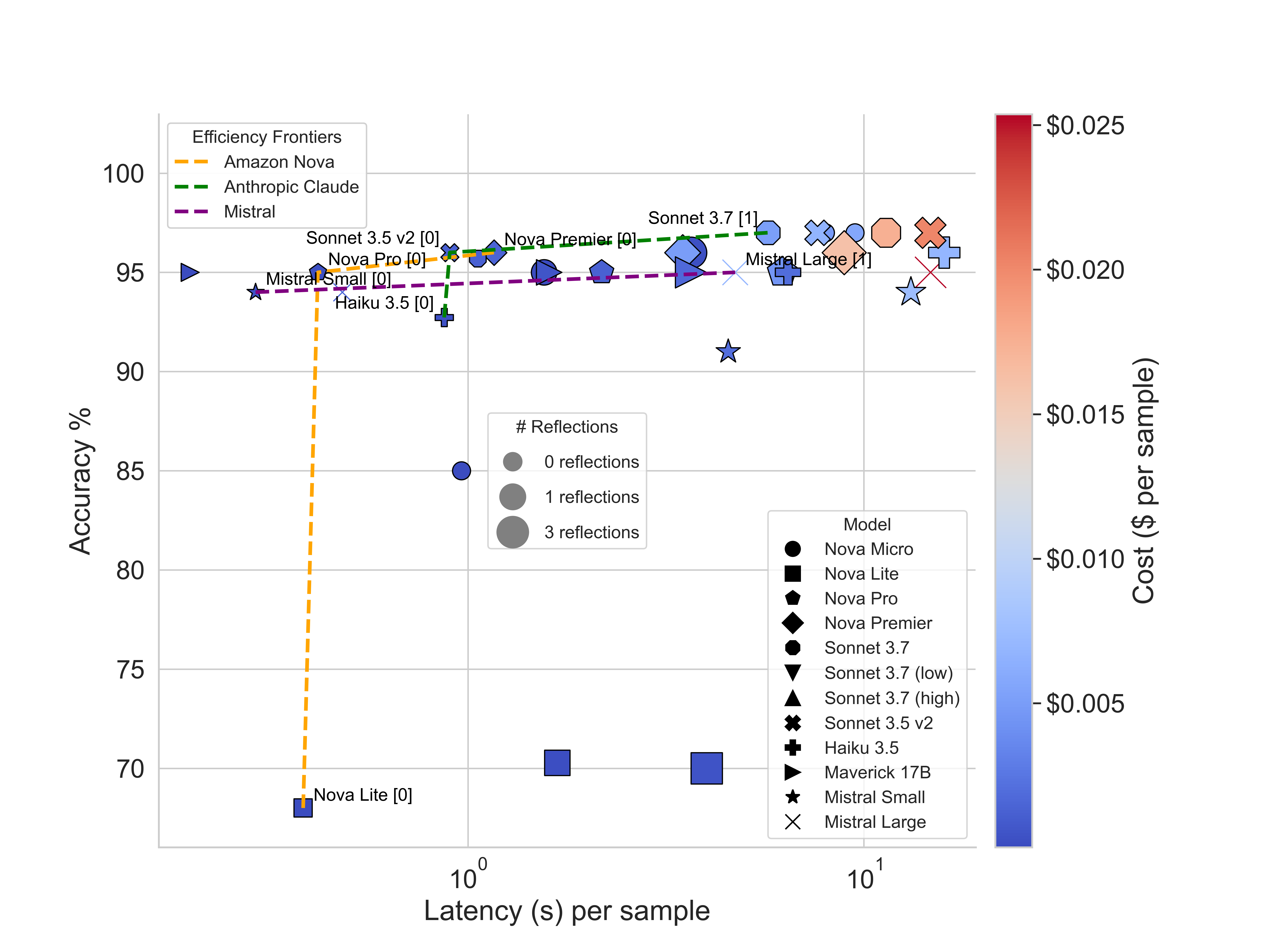}
    \caption{Accuracy-Latency Pareto Frontiers}
\end{subfigure}
\caption{Inference-Time Performance (IMDB)}
\label{fig:imdb}
\end{figure}

On sentiment analysis, Figure \ref{fig:imdb}(a) clearly illustrates the positive impact of self-reflection on the accuracy across the LLMs. For the most models, adding reflection rounds improves accuracy, though with diminishing returns after the 1st reflection. Nova Micro shows the highest relative improvement from 0 to 1 reflections (85\% to 95\% accuracy), while maintaining similar latency to 0 reflections of Sonnet 3.7 (1.56 vs 1.06) and similar resulting accuracy (95\% vs 95.7\%) at 1/18th of the cost. Nova Pro, Premier and Llama Maverick are the only models whose accuracy is not affected by reflection. There is an outlier, Mistral Small, which decreases in accuracy with reflections.

As depicted on Figure \ref{fig:imdb}(b), for applications requiring the highest possible accuracy, Sonnet 3.5 without reflection or Sonnet 3.7 with 1 reflection round offer the best performance. Built-in reasoning in Claude 3.7 performs similar to 1 round of self-reflection regardless of the thinking budget, but introduces higher latency and cost, making these configurations less attractive. For cost-sensitive deployments with moderate latency requirements, Nova Premier with 0 reflections presents a good compromise. Interestingly, Nova Micro with 3 reflections is able to reach a higher accuracy compared to Nova Premier, but results in a substantially higher overall latency and a marginally higher cost.

\vspace*{-1ex}

The results indicate that despite the ambiguity of the sentiment analysis task, LLMs consistently benefit from self-reflection in this domain. At the same time, the average gains are one order of magnitude smaller compared to mathematical reasoning, which makes inference-time techniques less attractive in terms of the cost-latency implications that may outweigh the accuracy gains.

\subsection{Translation (Flores-200)}

\begin{figure}[H]
\begin{subfigure}[b]{0.45\textwidth}
    \includegraphics[width=\textwidth,trim={0 0.25cm 0 0},clip]{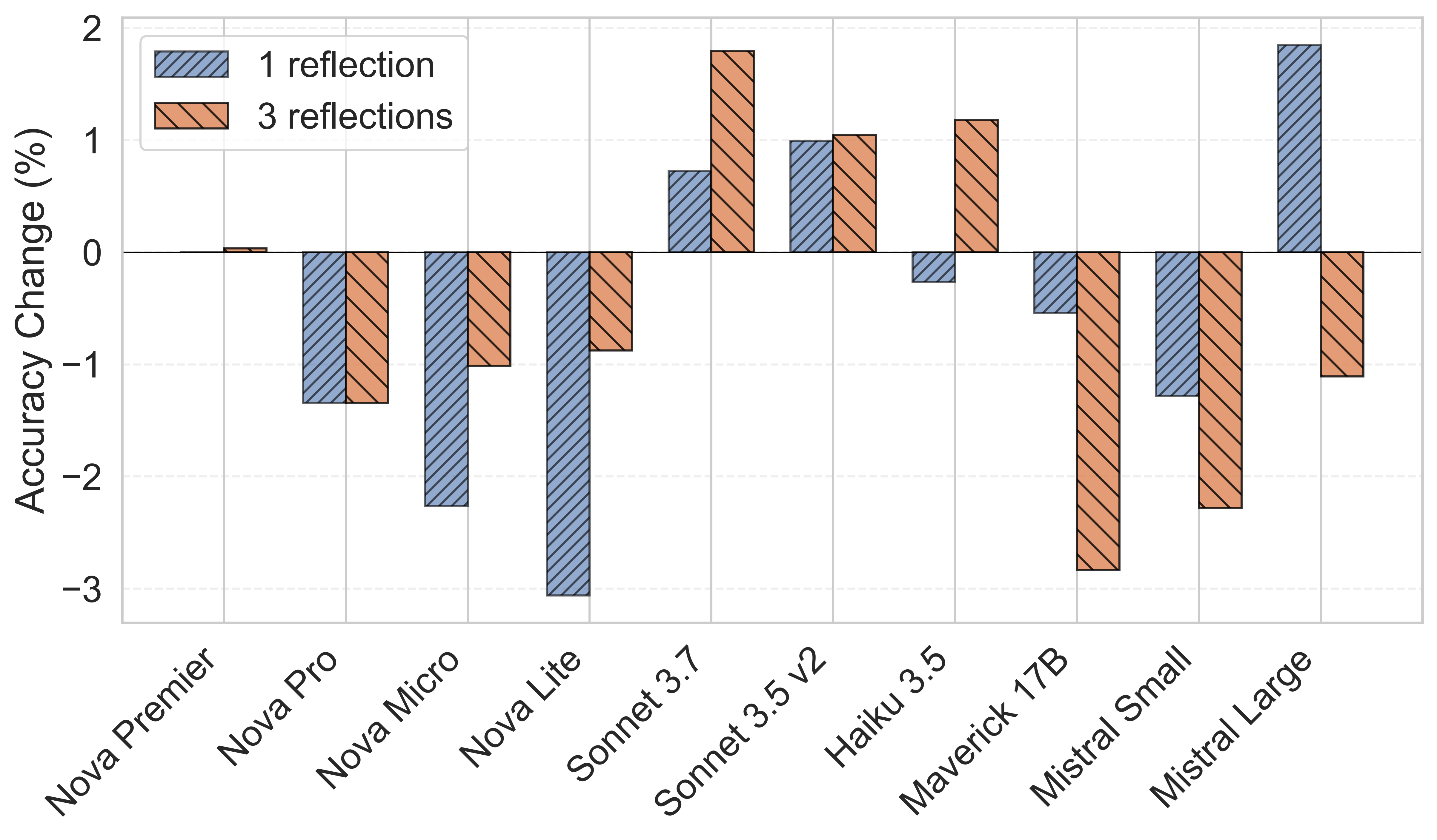}
    \caption{Relative Self-Reflection Gains}
\end{subfigure}
\hfill
\begin{subfigure}[b]{0.45\textwidth}
    \centering
    \includegraphics[width=\textwidth,trim={0 0.25cm 0 0},clip]{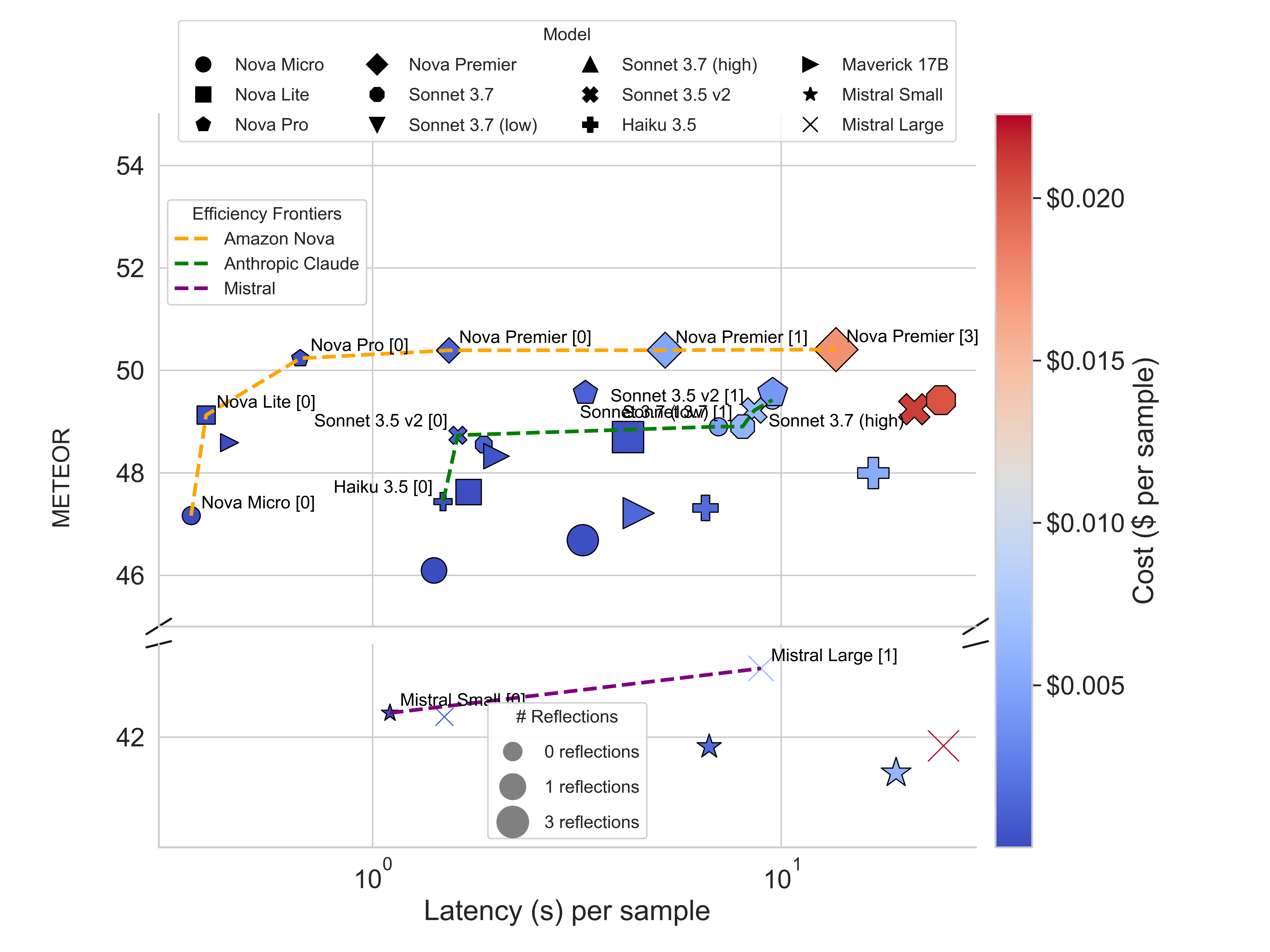}
    \caption{Accuracy-Latency Pareto Frontiers}
\end{subfigure}
\caption{Inference-Time Performance (Flores-200)}
\label{fig:flores}
\end{figure}

Figure \ref{fig:flores}(a) highlights distinct divergence in translation performance across LLM families. Claude models generally demonstrate enhanced performance after reflection. In contrast, all Amazon Nova models except Nova Premier exhibit an inverse trend, where incorporating 1 reflection diminishes translation accuracy. Despite some recovery when going from 1 to 3 reflections, Nova Micro, Lite and Pro still under-perform compared to their baseline with no reflection. Mistral Small and Llama Maverick also show negative performance with 1 reflection but unlike Amazon Nova, there is no recovery after 3. Mistral Large shows initial improvement after 1 reflection round but after 3 reflections we see a degradation of similar scale to the Nova models.  his implies that for Mistral Large only 1 reflection would be recommended where as using self-reflection for Amazon Nova LLMs, Mistral Small and Llama Maverick in translation tasks is not recommended.

\vspace*{-1ex}

Figure \ref{fig:flores}(b) suggests that Amazon Nova models dominate all considered Claude models in the latency-accuracy space. Nova Pro reaches higher accuracy when all Claude variants at a lower latency compared to Haiku 3.5. Furthermore, Nova Premier with self-reflection provides a further marginal gain in translation accuracy, but brings a substantial latency increase. This suggests that Amazon Nova performs particularly well in translation tasks, with an important caveat that integrating self-reflection for smaller variants hurts their performance. Focusing on Claude family, we note that Sonnet 3.7 built-in reasoning with a high thinking budget achieves the best METEOR score among Claude models, outperforming low thinking budget, self-reflections, and other Claude variants.
\subsection{Ablation Studies}
\label{sec:ablations}

{\textbf{Reflection Transitions}}

\begin{figure}[h]\centering
\begin{subfigure}[b]{0.43\textwidth}
    \centering
    \includegraphics[width=\textwidth,trim={1.5cm 0 0 0.5cm},clip]{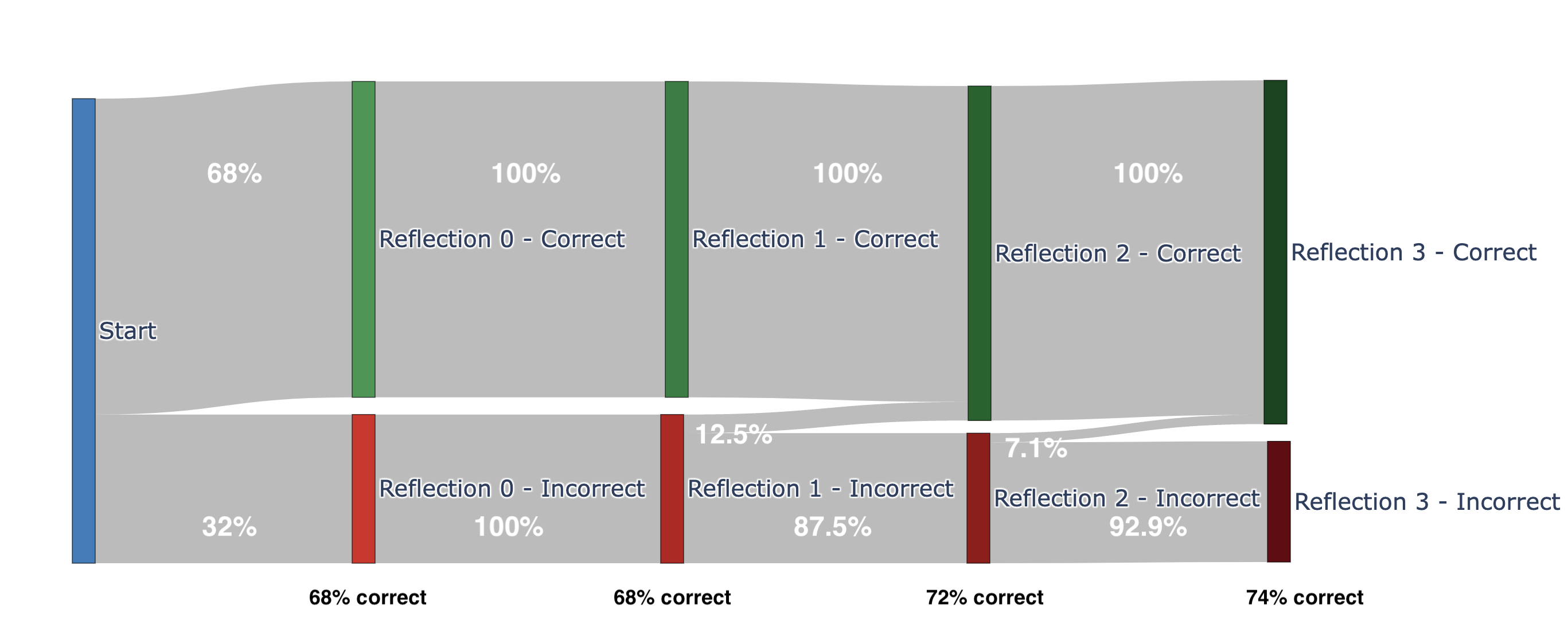}
    \caption{Claude Sonnet 3.5 v2}
\end{subfigure}
\begin{subfigure}[b]{0.43\textwidth}
    \centering
    \includegraphics[width=\textwidth,trim={1.5cm 0 0 0cm},clip]{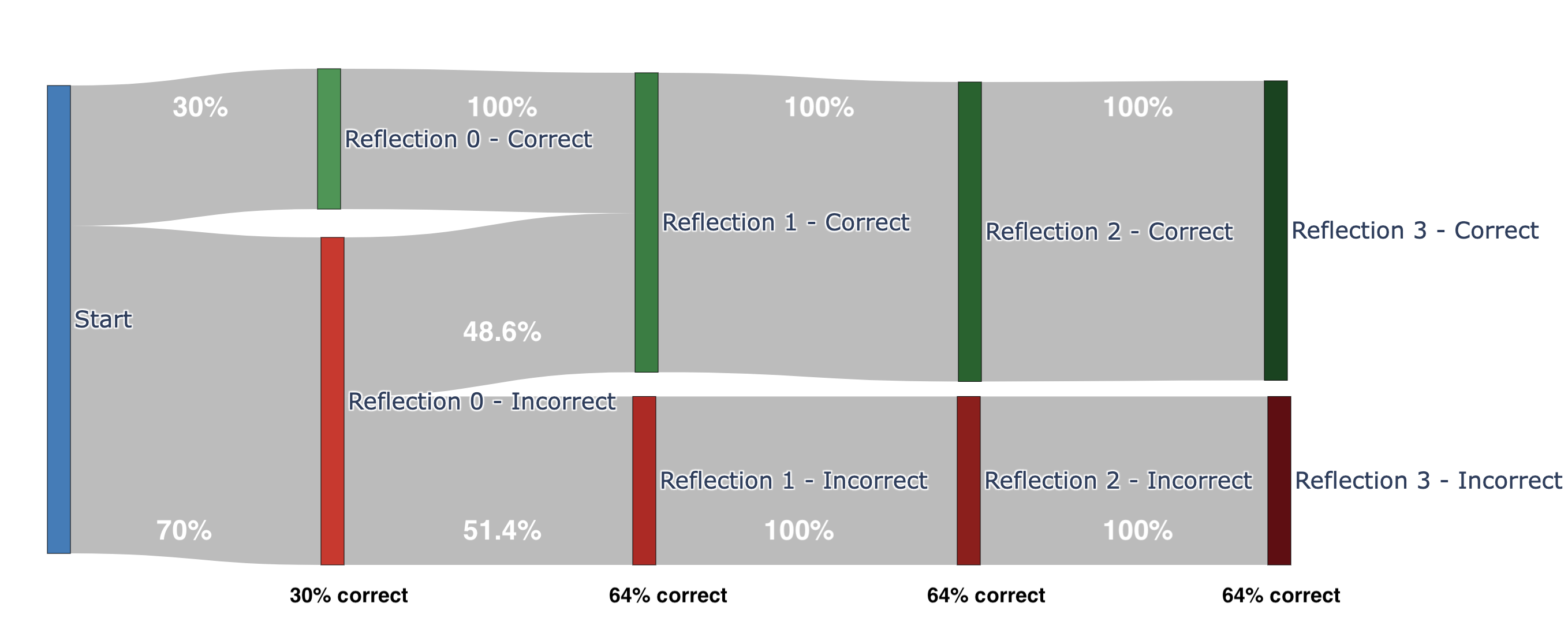}
    \caption{Amazon Nova Micro}
\end{subfigure}
\caption{Errors Across Reflections (Math500)}
\label{fig:math500-sankey}
\end{figure}

\noindent
Figure \ref{fig:math500-sankey} illustrates how LLM performance evolves throughout self-reflection rounds. We focus on mathematical reasoning, which benefits most from reflection and showcase results for 2 LLMs from model families with distinct performance patterns: Claude Sonnet 3.5 and Nova Micro. 

\vspace*{-1ex}

Sonnet 3.5 v2 demonstrates superior initial accuracy at 68\% compared to Nova Micro's 30\%. Through 3 reflection stages, Sonnet 3.5 shows consistent improvements, ultimately reaching 74\% accuracy. Interestingly, while the first reflection does not change Sonnet's accuracy, each subsequent round successfully corrects a portion of incorrect responses. In contrast, Nova Micro exhibits a dramatic improvement during the first reflection, jumping to 64\% accuracy after successfully correcting 48.6\% of initial errors. However, Nova's performance plateaus thereafter, showing no further improvement in subsequent reflection stages.
Another notable pattern across both LLMs is their perfect preservation of initially correct responses throughout all reflection rounds. These findings suggest that smaller models like Nova Micro have capacity for initial self-correction, whereas more capable LLMs like Sonnet 3.5 have stronger foundational performance and potential for continuous improvement through iterative reflection rounds.

\begin{table*}[t]\centering    
\caption{Impact of Feedback Mechanisms on Self-Reflection}
\label{tab:verifiers}
\begin{tabular}{@{\extracolsep{6pt}} l|cc|cc|cc}
    \hline
    & \multicolumn{2}{c|}{No feedback} & \multicolumn{2}{c|}{LLM judge feedback} & \multicolumn{2}{c}{SQL execution feedback} \\
    \cline{2-7}
    \multirow{-2}{*}{Model} & \multicolumn{1}{c}{1 round} & \multicolumn{1}{c|}{3 rounds} & \multicolumn{1}{c}{1 round} & \multicolumn{1}{c|}{3 rounds} & \multicolumn{1}{c}{1 round} & \multicolumn{1}{c}{3 rounds} \\
    \hline
    Amazon Nova Premier  & 72.58 & \textbf{74.98} & 73.97 & 72.58 & 73.74 & 71.14 \\
    Amazon Nova Pro      & 71.75 & \textbf{73.67} & 71.71 & 66.96 & 68.62 & 73.50 \\
    Amazon Nova Lite     & 75.41 & 73.05 & \textbf{79.57} & 74.02 & 72.63 & 72.83 \\
    Amazon Nova Micro    & 70.73 & 72.14 & \textbf{77.34} & 75.77 & 73.15 & 70.41 \\
    \hline
    Claude Sonnet 3.7    & 70.78 & 72.69 & 70.82 & 66.78 & 67.20 & \textbf{73.32} \\
    Claude Sonnet 3.5 v2 & 65.71 & 64.99 & 67.28 & 65.43 & 67.22 & \textbf{67.33} \\
    Claude Haiku 3.5     & 67.09 & 66.36 & 68.16 & 68.64 & 68.56 & \textbf{72.58} \\
    \hline
\end{tabular}
\vspace{0.15cm}
\end{table*}

\vspace{1ex}

\noindent
{\textbf{Reflection Feedback}}

\noindent
Table \ref{tab:verifiers} investigates if providing informative feedback to the LLM between self-reflection rounds facilitates stronger accuracy gains. We focus on text-to-SQL and compare 2 feedback mechanisms as LLM context: i) output of SQL query execution; ii) LLM-as-a-judge response with Nova Pro judge. 

The results reveal no clearly dominating feedback strategy. On average, incorporating feedback mechanisms improves reflection quality in 61\% of cases, confirming that additional feedback can be beneficial. However, model families respond differently to feedback types: Amazon Nova generally performs better with LLM-as-judge feedback or no feedback at all, while Claude shows higher accuracy with SQL execution feedback. This may be explained by the fact that Amazon Nova Pro judge is not able to provide stronger feedback to Claude models compared to their own reasoning, which may risk getting them off track. These findings emphasise the importance of identifying optimal configurations for specific business applications  by accounting for resource constraints, the LLM being used, and the task domain. Throughout our experiments, we consistently find that no single inference optimisation strategy proves universally effective across the diverse range of scenarios we tested.

\section{Results on Real-World Deployment}
\label{sec:case_study}

To validate our benchmark findings in a production environment, we present the real-world evaluation of a self-reflection-enhanced marketing content localisation system at Lounge by Zalando. Zalando is an online multi-brand fashion destination with more than 52 million active customers. Lounge by Zalando represents a shopping club, where customers can browse through a curated selection of fashion products. One of the critical business tasks at Lounge By Zalando is localising the marketing content for different European markets, including different distribution channels such as email newsletters, push notifications, and display.

Prior to our deployment, manual localisation process at Lounge by Zalando created significant bottlenecks, as copywriters required 2-3 days per campaign to localise content for major markets, limiting campaign agility and time-to-market. The core challenge was not merely translation into a local language, but sophisticated localisation incorporating: (i) market-specific tonality guidelines (e.g. using formal or informal pronouns when referring to a customer, (ii) local regulatory compliance (e.g. using correct terms for different sales types), and (iii) consistent brand voice adaptation.

\subsection{Technical Performance Evaluation}

To perform initial evaluation, we collect ground truth localised texts produced by copywriters independently from our system. We compare generated and human localisations using three metrics:

\begin{itemize}
    \item BLEU score comparing the generated and ground truth localisations;
    \item METEOR score comparing the generated and ground truth localisations;
    \item LLM-as-a-judge score. The judge (Claude Sonnet 3.5 in our case) picks the best localisation out of the generated and ground truth versions. We then aggregate preferences across the dataset to calculate the judge's preferred generation or whethet there was a tie.
\end{itemize}

\begin{table*}[t]\centering    
\caption{Self-Reflection Performance on Real-World Marketing Content Localisation Task}
\label{tab:zalando}
\begin{tabular}{@{\extracolsep{6pt}} l|ccc|ccc}
    \hline
    & \multicolumn{3}{c|}{No reflection} & \multicolumn{3}{c}{Self-reflection with LLM judge feedback} \\
    \cline{2-7}
    \multirow{-2}{*}{Language} & \multicolumn{1}{c}{BLEU} & \multicolumn{1}{c}{METEOR} & \multicolumn{1}{c|}{LLM judge score} & \multicolumn{1}{c}{BLEU} & \multicolumn{1}{c}{METEOR} & \multicolumn{1}{c}{LLM judge score} \\
    \hline
    French  & \textbf{0.16} & \textbf{0.47} & 0.61 & 0.14 & 0.42 & \textbf{0.62} \\
    Spanish & 0.29 & \textbf{0.61} & 0.49 & \textbf{0.29} & 0.59 & \textbf{0.50} \\
    German  & 0.32 & 0.61 & 0.38 & \textbf{0.33} & \textbf{0.62} & \textbf{0.47} \\
    \hline
\end{tabular}
\end{table*}

Table \ref{tab:zalando} illustrates the results on three markets. Considering the LLM-as-a-judge metric, we see that self-reflection improves the localisation quality on all languages, with the strongest gains observed for German. Here, the number of cases where the generated localisation is better than or the same quality as the human translation increases from 38\% to 47\%. On French and Spanish markets, the no-reflection version demonstrates very strong performance with 61\% and 49\%, diminishing the gains from self-reflection. 

Considering the text similarity metrics BLEU and METEOR, we observe mixed results with consistent improvements from self-reflection on the German market, its negative impact on localisation quality on the French market and similar performance with and without reflection on the Spanish market. It is important to note that manual inspection of selected localisations indicated that LLM-as-a-judge provides a more reliable quality measure, as similarity metrics do not incorporate the tonality guidelines and do not account for multiple accepted alternatives of formulating a sentence.

\subsection{Human Expert Evaluation}

To complement our automated metrics, we conduct a blind A/B evaluation with domain expert copywriters. Three human experts, each with multiple years of experience in marketing localisation for European markets, were asked to evaluate localised content without knowledge of which leveraged self-reflection. The experts assessed localisations across the same five evaluation criteria and reported the list of issues violating the guidelines. The results provided in Table \ref{tab:expert_eval} reveal significant improvements from self-reflection.

Expert evaluation results demonstrate substantial improvements in content quality, particularly for French localisations where self-reflection reduced identified issues by 88\% (from 384 to 46 issues). Spanish localisations showed a 39\% reduction in issue, while for German 100\% of original 15 issues were resolved. These results are in line with LLM-as-a-judge scores, which also favor the localisation variants with self-reflection.
These findings underscore why business metrics are essential for evaluating production NLP systems. Technical metrics provide relevant development feedback, but real world metrics such as domain expert evaluations results in higher time-to-market acceleration, cost reductions, and compliance adherence.

\clearpage

Overall, the results indicate it is valuable to employ self-reflection on markets with more challenging localisation rules (such as German), whereas using it on markets where the base model already achieves high quality provides minimal quality gains that may not justify the additional LLM cost. While this confirms some of the findings from Sections \ref{sec_inputs} and \ref{sec:ablations}, it also emphasises the importance of testing self-reflection performance on a specific dataset, as the gains may vary significantly depending on the use case.

Given that self-reflection enhances output quality at increased computational cost, we examine whether prompt caching \cite{gim2024promptcachemodularattention} can partly mitigate this cost overhead. Our analysis demonstrates that combining self-reflection with prompt caching achieves cost reductions of up to 28\% when employing three reflection rounds with detailed cost-latency analysis presented in Appendix B.4.

\begin{table}[h]\centering    
\caption{Human Expert Evaluation Results}
\label{tab:expert_eval}
\begin{tabular}{@{\extracolsep{0pt}} l|cc}
    \hline
    Language & No reflection & Self-reflection \\
    \hline
    French & 384 issues & \textbf{46 issues} (-88\%) \\
    Spanish & 49 issues & \textbf{30 issues} (-39\%) \\
    German & 15 issues & \textbf{0 issues} (-100\%) \\
    \hline
\end{tabular}
\end{table}
%
%

\section{Conclusion}
\label{sec_conclusion}

This paper presents a systematic analysis of inference optimisation techniques such as self-reflection and budget tuning across different domains, base models, and reflection parameters. We derive Pareto frontiers in accuracy-latency space, test self-reflection in a production deployment, and provide actionable recommendations to practitioners regarding suitable inference optimisation methods for real-world applications.

Our results reveal no universally dominant inference strategy, with both the magnitude and direction of performance impacts varying significantly across tasks. Self-reflection consistently improves performance in math (with gains up to 220\%) and sentiment analysis, while showing mixed or negative effects in translation and text-to-SQL generation. This domain-specific variability highlights the importance of empirical evaluation before deploying inference optimisation techniques in production.

Several key patterns emerge from our analysis: (i) smaller LLMs often benefit more dramatically from reflection than larger ones; (ii) a single reflection round frequently captures most potential performance benefits; (iii) in several cases, smaller LLMs with reflection outperform larger models without it, offering potential cost savings; and (iv) Claude's built-in reasoning sometimes underperforms compared to self-reflection techniques despite its specialised design, and results in higher additional cost as it does not benefit from prompt caching.

For practitioners, these findings suggest task-specific optimisation strategies. For math, self-reflection is highly recommended, with Amazon Nova Micro offering an excellent cost-performance balance. For text-to-SQL, Amazon Nova generally outperform Claude variants, with minimal reflection recommended. For sentiment analysis, most models benefit from reflection, though gains may not justify increased costs. For translation, Claude generally benefits from reflection while Amazon Nova performs better without it. The case study on real-world marketing content localisation data confirms that self-reflection gains with Claude models may be higher on the tasks that are more challenging. 

In future work, we aim to conduct a deeper interpretative analysis of the budget tuning methods, including providing transition analysis of the generated thinking tokens. We also wish to expand our analysis outside of the Amazon Nova, Mistral and Anthropic Claude model families to understand the influence of inference-time compute techniques on other leading model providers. We would also want to understand the benefits of combining complimentary techniques from inference-time compute such as parallel sampling, best-of-N majority voting and others. 


%
%


\clearpage
\bibliographystyle{ACM-Reference-Format}
\bibliography{references}


\begin{thebibliography}{26}


\ifx \showCODEN    \undefined \def \showCODEN     #1{\unskip}     \fi
\ifx \showDOI      \undefined \def \showDOI       #1{#1}\fi
\ifx \showISBNx    \undefined \def \showISBNx     #1{\unskip}     \fi
\ifx \showISBNxiii \undefined \def \showISBNxiii  #1{\unskip}     \fi
\ifx \showISSN     \undefined \def \showISSN      #1{\unskip}     \fi
\ifx \showLCCN     \undefined \def \showLCCN      #1{\unskip}     \fi
\ifx \shownote     \undefined \def \shownote      #1{#1}          \fi
\ifx \showarticletitle \undefined \def \showarticletitle #1{#1}   \fi
\ifx \showURL      \undefined \def \showURL       {\relax}        \fi
\providecommand\bibfield[2]{#2}
\providecommand\bibinfo[2]{#2}
\providecommand\natexlab[1]{#1}
\providecommand\showeprint[2][]{arXiv:#2}

\bibitem[Akyürek et~al\mbox{.}(2025)]%
        {akyürek2025surprisingeffectivenesstesttimetraining}
\bibfield{author}{\bibinfo{person}{Ekin Akyürek}, \bibinfo{person}{Mehul Damani}, \bibinfo{person}{Adam Zweiger}, \bibinfo{person}{Linlu Qiu}, \bibinfo{person}{Han Guo}, \bibinfo{person}{Jyothish Pari}, \bibinfo{person}{Yoon Kim}, {and} \bibinfo{person}{Jacob Andreas}.} \bibinfo{year}{2025}\natexlab{}.
\newblock \bibinfo{title}{The Surprising Effectiveness of Test-Time Training for Few-Shot Learning}.
\newblock
\newblock
\showeprint[arxiv]{2411.07279}~[cs.AI]
\urldef\tempurl%
\url{https://arxiv.org/abs/2411.07279}
\showURL{%
\tempurl}


\bibitem[Anthropic(2024)]%
        {claude3}
\bibfield{author}{\bibinfo{person}{Anthropic}.} \bibinfo{year}{2024}\natexlab{}.
\newblock \showarticletitle{{The Claude 3 Model Family: Opus, Sonnet, Haiku}}.
\newblock  (\bibinfo{year}{2024}).
\newblock


\bibitem[Besta et~al\mbox{.}(2024)]%
        {Besta_2024}
\bibfield{author}{\bibinfo{person}{Maciej Besta}, \bibinfo{person}{Nils Blach}, \bibinfo{person}{Ales Kubicek}, \bibinfo{person}{Robert Gerstenberger}, \bibinfo{person}{Michal Podstawski}, \bibinfo{person}{Lukas Gianinazzi}, \bibinfo{person}{Joanna Gajda}, \bibinfo{person}{Tomasz Lehmann}, \bibinfo{person}{Hubert Niewiadomski}, \bibinfo{person}{Piotr Nyczyk}, {and} \bibinfo{person}{Torsten Hoefler}.} \bibinfo{year}{2024}\natexlab{}.
\newblock \showarticletitle{Graph of Thoughts: Solving Elaborate Problems with Large Language Models}.
\newblock \bibinfo{journal}{\emph{Proceedings of the AAAI Conference on Artificial Intelligence}} \bibinfo{volume}{38}, \bibinfo{number}{16} (\bibinfo{date}{March} \bibinfo{year}{2024}), \bibinfo{pages}{17682–17690}.
\newblock
\showISSN{2159-5399}
\urldef\tempurl%
\url{https://doi.org/10.1609/aaai.v38i16.29720}
\showDOI{\tempurl}


\bibitem[Chen et~al\mbox{.}(2024)]%
        {chen2024teaching}
\bibfield{author}{\bibinfo{person}{Xinyun Chen}, \bibinfo{person}{Maxwell Lin}, \bibinfo{person}{Nathanael Sch{\"a}rli}, {and} \bibinfo{person}{Denny Zhou}.} \bibinfo{year}{2024}\natexlab{}.
\newblock \showarticletitle{Teaching Large Language Models to Self-Debug}. In \bibinfo{booktitle}{\emph{The Twelfth International Conference on Learning Representations}}.
\newblock
\urldef\tempurl%
\url{https://openreview.net/forum?id=KuPixIqPiq}
\showURL{%
\tempurl}


\bibitem[Chiang et~al\mbox{.}(2024)]%
        {chiang2024chatbotarenaopenplatform}
\bibfield{author}{\bibinfo{person}{Wei-Lin Chiang}, \bibinfo{person}{Lianmin Zheng}, \bibinfo{person}{Ying Sheng}, \bibinfo{person}{Anastasios~Nikolas Angelopoulos}, \bibinfo{person}{Tianle Li}, \bibinfo{person}{Dacheng Li}, \bibinfo{person}{Hao Zhang}, \bibinfo{person}{Banghua Zhu}, \bibinfo{person}{Michael Jordan}, \bibinfo{person}{Joseph~E. Gonzalez}, {and} \bibinfo{person}{Ion Stoica}.} \bibinfo{year}{2024}\natexlab{}.
\newblock \bibinfo{title}{Chatbot Arena: An Open Platform for Evaluating LLMs by Human Preference}.
\newblock
\newblock
\showeprint[arxiv]{2403.04132}~[cs.AI]
\urldef\tempurl%
\url{https://arxiv.org/abs/2403.04132}
\showURL{%
\tempurl}


\bibitem[Costa-Juss{\`a} et~al\mbox{.}(2022)]%
        {nllb2022}
\bibfield{author}{\bibinfo{person}{Marta~R Costa-Juss{\`a}}, \bibinfo{person}{James Cross}, \bibinfo{person}{Onur {\c{C}}elebi}, \bibinfo{person}{Maha Elbayad}, \bibinfo{person}{Kenneth Heafield}, \bibinfo{person}{Kevin Heffernan}, \bibinfo{person}{Elahe Kalbassi}, \bibinfo{person}{Janice Lam}, \bibinfo{person}{Daniel Licht}, \bibinfo{person}{Jean Maillard}, {et~al\mbox{.}}} \bibinfo{year}{2022}\natexlab{}.
\newblock \showarticletitle{No language left behind: Scaling human-centered machine translation}.
\newblock \bibinfo{journal}{\emph{arXiv preprint arXiv:2207.04672}} (\bibinfo{year}{2022}).
\newblock


\bibitem[Gim et~al\mbox{.}(2024)]%
        {gim2024promptcachemodularattention}
\bibfield{author}{\bibinfo{person}{In Gim}, \bibinfo{person}{Guojun Chen}, \bibinfo{person}{Seung seob Lee}, \bibinfo{person}{Nikhil Sarda}, \bibinfo{person}{Anurag Khandelwal}, {and} \bibinfo{person}{Lin Zhong}.} \bibinfo{year}{2024}\natexlab{}.
\newblock \bibinfo{title}{Prompt Cache: Modular Attention Reuse for Low-Latency Inference}.
\newblock
\newblock
\showeprint[arxiv]{2311.04934}~[cs.CL]
\urldef\tempurl%
\url{https://arxiv.org/abs/2311.04934}
\showURL{%
\tempurl}


\bibitem[Hoffmann et~al\mbox{.}(2022)]%
        {hoffmann2022trainingcomputeoptimallargelanguage}
\bibfield{author}{\bibinfo{person}{Jordan Hoffmann}, \bibinfo{person}{Sebastian Borgeaud}, \bibinfo{person}{Arthur Mensch}, \bibinfo{person}{Elena Buchatskaya}, \bibinfo{person}{Trevor Cai}, \bibinfo{person}{Eliza Rutherford}, \bibinfo{person}{Diego de Las~Casas}, \bibinfo{person}{Lisa~Anne Hendricks}, \bibinfo{person}{Johannes Welbl}, \bibinfo{person}{Aidan Clark}, \bibinfo{person}{Tom Hennigan}, \bibinfo{person}{Eric Noland}, \bibinfo{person}{Katie Millican}, \bibinfo{person}{George van~den Driessche}, \bibinfo{person}{Bogdan Damoc}, \bibinfo{person}{Aurelia Guy}, \bibinfo{person}{Simon Osindero}, \bibinfo{person}{Karen Simonyan}, \bibinfo{person}{Erich Elsen}, \bibinfo{person}{Jack~W. Rae}, \bibinfo{person}{Oriol Vinyals}, {and} \bibinfo{person}{Laurent Sifre}.} \bibinfo{year}{2022}\natexlab{}.
\newblock \bibinfo{title}{Training Compute-Optimal Large Language Models}.
\newblock
\newblock
\showeprint[arxiv]{2203.15556}~[cs.CL]
\urldef\tempurl%
\url{https://arxiv.org/abs/2203.15556}
\showURL{%
\tempurl}


\bibitem[H{\"u}botter et~al\mbox{.}(2025)]%
        {hubotter2025efficiently}
\bibfield{author}{\bibinfo{person}{Jonas H{\"u}botter}, \bibinfo{person}{Sascha Bongni}, \bibinfo{person}{Ido Hakimi}, {and} \bibinfo{person}{Andreas Krause}.} \bibinfo{year}{2025}\natexlab{}.
\newblock \showarticletitle{Efficiently Learning at Test-Time: Active Fine-Tuning of {LLM}s}. In \bibinfo{booktitle}{\emph{The Thirteenth International Conference on Learning Representations}}.
\newblock
\urldef\tempurl%
\url{https://openreview.net/forum?id=NS1G1Uhny3}
\showURL{%
\tempurl}


\bibitem[Intelligence(2024)]%
        {Intelligence2024}
\bibfield{author}{\bibinfo{person}{Amazon Artificial~General Intelligence}.} \bibinfo{year}{2024}\natexlab{}.
\newblock \showarticletitle{The Amazon Nova family of models: Technical report and model card}.
\newblock \bibinfo{journal}{\emph{Amazon Technical Reports}} (\bibinfo{year}{2024}).
\newblock
\urldef\tempurl%
\url{https://www.amazon.science/publications/the-amazon-nova-family-of-models-technical-report-and-model-card}
\showURL{%
\tempurl}


\bibitem[Lavie and Agarwal(2007)]%
        {meteor}
\bibfield{author}{\bibinfo{person}{Alon Lavie} {and} \bibinfo{person}{Abhaya Agarwal}.} \bibinfo{year}{2007}\natexlab{}.
\newblock \showarticletitle{Meteor: an automatic metric for MT evaluation with high levels of correlation with human judgments}. In \bibinfo{booktitle}{\emph{Proceedings of the Second Workshop on Statistical Machine Translation}} (Prague, Czech Republic) \emph{(\bibinfo{series}{StatMT '07})}. \bibinfo{publisher}{Association for Computational Linguistics}, \bibinfo{address}{USA}, \bibinfo{pages}{228–231}.
\newblock


\bibitem[Lee et~al\mbox{.}(2024)]%
        {lee2024rlaif}
\bibfield{author}{\bibinfo{person}{Harrison Lee}, \bibinfo{person}{Samrat Phatale}, \bibinfo{person}{Hassan Mansoor}, \bibinfo{person}{Kellie~Ren Lu}, \bibinfo{person}{Thomas Mesnard}, \bibinfo{person}{Johan Ferret}, \bibinfo{person}{Colton Bishop}, \bibinfo{person}{Ethan Hall}, \bibinfo{person}{Victor Carbune}, {and} \bibinfo{person}{Abhinav Rastogi}.} \bibinfo{year}{2024}\natexlab{}.
\newblock \bibinfo{title}{{RLAIF}: Scaling Reinforcement Learning from Human Feedback with {AI} Feedback}.
\newblock
\newblock
\urldef\tempurl%
\url{https://openreview.net/forum?id=AAxIs3D2ZZ}
\showURL{%
\tempurl}


\bibitem[Liang et~al\mbox{.}(2023)]%
        {liang2023holisticevaluationlanguagemodels}
\bibfield{author}{\bibinfo{person}{Percy Liang}, \bibinfo{person}{Rishi Bommasani}, \bibinfo{person}{Tony Lee}, \bibinfo{person}{Dimitris Tsipras}, \bibinfo{person}{Dilara Soylu}, \bibinfo{person}{Michihiro Yasunaga}, \bibinfo{person}{Yian Zhang}, \bibinfo{person}{Deepak Narayanan}, \bibinfo{person}{Yuhuai Wu}, \bibinfo{person}{Ananya Kumar}, \bibinfo{person}{Benjamin Newman}, \bibinfo{person}{Binhang Yuan}, \bibinfo{person}{Bobby Yan}, \bibinfo{person}{Ce Zhang}, \bibinfo{person}{Christian Cosgrove}, \bibinfo{person}{Christopher~D. Manning}, \bibinfo{person}{Christopher Ré}, \bibinfo{person}{Diana Acosta-Navas}, \bibinfo{person}{Drew~A. Hudson}, \bibinfo{person}{Eric Zelikman}, \bibinfo{person}{Esin Durmus}, \bibinfo{person}{Faisal Ladhak}, \bibinfo{person}{Frieda Rong}, \bibinfo{person}{Hongyu Ren}, \bibinfo{person}{Huaxiu Yao}, \bibinfo{person}{Jue Wang}, \bibinfo{person}{Keshav Santhanam}, \bibinfo{person}{Laurel Orr}, \bibinfo{person}{Lucia Zheng}, \bibinfo{person}{Mert Yuksekgonul},
  \bibinfo{person}{Mirac Suzgun}, \bibinfo{person}{Nathan Kim}, \bibinfo{person}{Neel Guha}, \bibinfo{person}{Niladri Chatterji}, \bibinfo{person}{Omar Khattab}, \bibinfo{person}{Peter Henderson}, \bibinfo{person}{Qian Huang}, \bibinfo{person}{Ryan Chi}, \bibinfo{person}{Sang~Michael Xie}, \bibinfo{person}{Shibani Santurkar}, \bibinfo{person}{Surya Ganguli}, \bibinfo{person}{Tatsunori Hashimoto}, \bibinfo{person}{Thomas Icard}, \bibinfo{person}{Tianyi Zhang}, \bibinfo{person}{Vishrav Chaudhary}, \bibinfo{person}{William Wang}, \bibinfo{person}{Xuechen Li}, \bibinfo{person}{Yifan Mai}, \bibinfo{person}{Yuhui Zhang}, {and} \bibinfo{person}{Yuta Koreeda}.} \bibinfo{year}{2023}\natexlab{}.
\newblock \bibinfo{title}{Holistic Evaluation of Language Models}.
\newblock
\newblock
\showeprint[arxiv]{2211.09110}~[cs.CL]
\urldef\tempurl%
\url{https://arxiv.org/abs/2211.09110}
\showURL{%
\tempurl}


\bibitem[Lightman et~al\mbox{.}(2023)]%
        {lightman2023letsverifystepstep}
\bibfield{author}{\bibinfo{person}{Hunter Lightman}, \bibinfo{person}{Vineet Kosaraju}, \bibinfo{person}{Yura Burda}, \bibinfo{person}{Harri Edwards}, \bibinfo{person}{Bowen Baker}, \bibinfo{person}{Teddy Lee}, \bibinfo{person}{Jan Leike}, \bibinfo{person}{John Schulman}, \bibinfo{person}{Ilya Sutskever}, {and} \bibinfo{person}{Karl Cobbe}.} \bibinfo{year}{2023}\natexlab{}.
\newblock \bibinfo{title}{Let's Verify Step by Step}.
\newblock
\newblock
\showeprint[arxiv]{2305.20050}~[cs.LG]
\urldef\tempurl%
\url{https://arxiv.org/abs/2305.20050}
\showURL{%
\tempurl}


\bibitem[Lightman et~al\mbox{.}(2024)]%
        {lightman2024lets}
\bibfield{author}{\bibinfo{person}{Hunter Lightman}, \bibinfo{person}{Vineet Kosaraju}, \bibinfo{person}{Yuri Burda}, \bibinfo{person}{Harrison Edwards}, \bibinfo{person}{Bowen Baker}, \bibinfo{person}{Teddy Lee}, \bibinfo{person}{Jan Leike}, \bibinfo{person}{John Schulman}, \bibinfo{person}{Ilya Sutskever}, {and} \bibinfo{person}{Karl Cobbe}.} \bibinfo{year}{2024}\natexlab{}.
\newblock \showarticletitle{Let's Verify Step by Step}. In \bibinfo{booktitle}{\emph{The Twelfth International Conference on Learning Representations}}.
\newblock
\urldef\tempurl%
\url{https://openreview.net/forum?id=v8L0pN6EOi}
\showURL{%
\tempurl}


\bibitem[Maas et~al\mbox{.}(2011)]%
        {maas-EtAl:2011:ACL-HLT2011}
\bibfield{author}{\bibinfo{person}{Andrew~L. Maas}, \bibinfo{person}{Raymond~E. Daly}, \bibinfo{person}{Peter~T. Pham}, \bibinfo{person}{Dan Huang}, \bibinfo{person}{Andrew~Y. Ng}, {and} \bibinfo{person}{Christopher Potts}.} \bibinfo{year}{2011}\natexlab{}.
\newblock \showarticletitle{Learning Word Vectors for Sentiment Analysis}. In \bibinfo{booktitle}{\emph{Proceedings of the 49th Annual Meeting of the Association for Computational Linguistics: Human Language Technologies}}. \bibinfo{publisher}{Association for Computational Linguistics}, \bibinfo{address}{Portland, Oregon, USA}, \bibinfo{pages}{142--150}.
\newblock
\urldef\tempurl%
\url{http://www.aclweb.org/anthology/P11-1015}
\showURL{%
\tempurl}


\bibitem[Madaan et~al\mbox{.}(2023)]%
        {madaan2023selfrefine}
\bibfield{author}{\bibinfo{person}{Aman Madaan}, \bibinfo{person}{Niket Tandon}, \bibinfo{person}{Prakhar Gupta}, \bibinfo{person}{Skyler Hallinan}, \bibinfo{person}{Luyu Gao}, \bibinfo{person}{Sarah Wiegreffe}, \bibinfo{person}{Uri Alon}, \bibinfo{person}{Nouha Dziri}, \bibinfo{person}{Shrimai Prabhumoye}, \bibinfo{person}{Yiming Yang}, \bibinfo{person}{Shashank Gupta}, \bibinfo{person}{Bodhisattwa~Prasad Majumder}, \bibinfo{person}{Katherine Hermann}, \bibinfo{person}{Sean Welleck}, \bibinfo{person}{Amir Yazdanbakhsh}, {and} \bibinfo{person}{Peter Clark}.} \bibinfo{year}{2023}\natexlab{}.
\newblock \showarticletitle{Self-Refine: Iterative Refinement with Self-Feedback}. In \bibinfo{booktitle}{\emph{Thirty-seventh Conference on Neural Information Processing Systems}}.
\newblock
\urldef\tempurl%
\url{https://openreview.net/forum?id=S37hOerQLB}
\showURL{%
\tempurl}


\bibitem[Meta(2025)]%
        {Llama4}
\bibfield{author}{\bibinfo{person}{Meta}.} \bibinfo{year}{2025}\natexlab{}.
\newblock \showarticletitle{The Llama 4 herd: The beginning of a new era of natively multimodal AI innovation}.
\newblock  (\bibinfo{year}{2025}).
\newblock


\bibitem[Meurer et~al\mbox{.}(2017)]%
        {sympy}
\bibfield{author}{\bibinfo{person}{Aaron Meurer}, \bibinfo{person}{Christopher~P. Smith}, \bibinfo{person}{Mateusz Paprocki}, \bibinfo{person}{Ond\v{r}ej \v{C}ert\'{i}k}, \bibinfo{person}{Sergey~B. Kirpichev}, \bibinfo{person}{Matthew Rocklin}, \bibinfo{person}{AMiT Kumar}, \bibinfo{person}{Sergiu Ivanov}, \bibinfo{person}{Jason~K. Moore}, \bibinfo{person}{Sartaj Singh}, \bibinfo{person}{Thilina Rathnayake}, \bibinfo{person}{Sean Vig}, \bibinfo{person}{Brian~E. Granger}, \bibinfo{person}{Richard~P. Muller}, \bibinfo{person}{Francesco Bonazzi}, \bibinfo{person}{Harsh Gupta}, \bibinfo{person}{Shivam Vats}, \bibinfo{person}{Fredrik Johansson}, \bibinfo{person}{Fabian Pedregosa}, \bibinfo{person}{Matthew~J. Curry}, \bibinfo{person}{Andy~R. Terrel}, \bibinfo{person}{\v{S}t\v{e}p\'{a}n Rou\v{c}ka}, \bibinfo{person}{Ashutosh Saboo}, \bibinfo{person}{Isuru Fernando}, \bibinfo{person}{Sumith Kulal}, \bibinfo{person}{Robert Cimrman}, {and} \bibinfo{person}{Anthony Scopatz}.} \bibinfo{year}{2017}\natexlab{}.
\newblock \showarticletitle{SymPy: symbolic computing in Python}.
\newblock \bibinfo{journal}{\emph{PeerJ Computer Science}}  \bibinfo{volume}{3} (\bibinfo{date}{Jan.} \bibinfo{year}{2017}), \bibinfo{pages}{e103}.
\newblock
\showISSN{2376-5992}
\urldef\tempurl%
\url{https://doi.org/10.7717/peerj-cs.103}
\showDOI{\tempurl}


\bibitem[Mistral(2024)]%
        {Mistral}
\bibfield{author}{\bibinfo{person}{Mistral}.} \bibinfo{year}{2024}\natexlab{}.
\newblock \showarticletitle{AI in abundance}.
\newblock  (\bibinfo{year}{2024}).
\newblock


\bibitem[Setlur et~al\mbox{.}(2025)]%
        {setlur2025rewarding}
\bibfield{author}{\bibinfo{person}{Amrith Setlur}, \bibinfo{person}{Chirag Nagpal}, \bibinfo{person}{Adam Fisch}, \bibinfo{person}{Xinyang Geng}, \bibinfo{person}{Jacob Eisenstein}, \bibinfo{person}{Rishabh Agarwal}, \bibinfo{person}{Alekh Agarwal}, \bibinfo{person}{Jonathan Berant}, {and} \bibinfo{person}{Aviral Kumar}.} \bibinfo{year}{2025}\natexlab{}.
\newblock \showarticletitle{Rewarding Progress: Scaling Automated Process Verifiers for {LLM} Reasoning}. In \bibinfo{booktitle}{\emph{The Thirteenth International Conference on Learning Representations}}.
\newblock
\urldef\tempurl%
\url{https://openreview.net/forum?id=A6Y7AqlzLW}
\showURL{%
\tempurl}


\bibitem[Snell et~al\mbox{.}(2025)]%
        {snell2025scaling}
\bibfield{author}{\bibinfo{person}{Charlie~Victor Snell}, \bibinfo{person}{Jaehoon Lee}, \bibinfo{person}{Kelvin Xu}, {and} \bibinfo{person}{Aviral Kumar}.} \bibinfo{year}{2025}\natexlab{}.
\newblock \showarticletitle{Scaling {LLM} Test-Time Compute Optimally Can be More Effective than Scaling Parameters for Reasoning}. In \bibinfo{booktitle}{\emph{The Thirteenth International Conference on Learning Representations}}.
\newblock
\urldef\tempurl%
\url{https://openreview.net/forum?id=4FWAwZtd2n}
\showURL{%
\tempurl}


\bibitem[Trung et~al\mbox{.}(2024)]%
        {trung-etal-2024-reft}
\bibfield{author}{\bibinfo{person}{Luong Trung}, \bibinfo{person}{Xinbo Zhang}, \bibinfo{person}{Zhanming Jie}, \bibinfo{person}{Peng Sun}, \bibinfo{person}{Xiaoran Jin}, {and} \bibinfo{person}{Hang Li}.} \bibinfo{year}{2024}\natexlab{}.
\newblock \showarticletitle{{R}e{FT}: Reasoning with Reinforced Fine-Tuning}. In \bibinfo{booktitle}{\emph{Proceedings of the 62nd Annual Meeting of the Association for Computational Linguistics (Volume 1: Long Papers)}}, \bibfield{editor}{\bibinfo{person}{Lun-Wei Ku}, \bibinfo{person}{Andre Martins}, {and} \bibinfo{person}{Vivek Srikumar}} (Eds.). \bibinfo{publisher}{Association for Computational Linguistics}, \bibinfo{address}{Bangkok, Thailand}, \bibinfo{pages}{7601--7614}.
\newblock
\urldef\tempurl%
\url{https://doi.org/10.18653/v1/2024.acl-long.410}
\showDOI{\tempurl}


\bibitem[Xi et~al\mbox{.}(2024)]%
        {r3}
\bibfield{author}{\bibinfo{person}{Zhiheng Xi}, \bibinfo{person}{Wenxiang Chen}, \bibinfo{person}{Boyang Hong}, \bibinfo{person}{Senjie Jin}, \bibinfo{person}{Rui Zheng}, \bibinfo{person}{Wei He}, \bibinfo{person}{Yiwen Ding}, \bibinfo{person}{Shichun Liu}, \bibinfo{person}{Xin Guo}, \bibinfo{person}{Junzhe Wang}, \bibinfo{person}{Honglin Guo}, \bibinfo{person}{Wei Shen}, \bibinfo{person}{Xiaoran Fan}, \bibinfo{person}{Yuhao Zhou}, \bibinfo{person}{Shihan Dou}, \bibinfo{person}{Xiao Wang}, \bibinfo{person}{Xinbo Zhang}, \bibinfo{person}{Peng Sun}, \bibinfo{person}{Tao Gui}, \bibinfo{person}{Qi Zhang}, {and} \bibinfo{person}{Xuanjing Huang}.} \bibinfo{year}{2024}\natexlab{}.
\newblock \showarticletitle{Training large language models for reasoning through reverse curriculum reinforcement learning}. In \bibinfo{booktitle}{\emph{Proceedings of the 41st International Conference on Machine Learning}} (Vienna, Austria) \emph{(\bibinfo{series}{ICML'24})}. \bibinfo{publisher}{JMLR.org}, Article \bibinfo{articleno}{2217}, \bibinfo{numpages}{19}~pages.
\newblock


\bibitem[Yao et~al\mbox{.}(2023)]%
        {yao2023treethoughtsdeliberateproblem}
\bibfield{author}{\bibinfo{person}{Shunyu Yao}, \bibinfo{person}{Dian Yu}, \bibinfo{person}{Jeffrey Zhao}, \bibinfo{person}{Izhak Shafran}, \bibinfo{person}{Thomas~L. Griffiths}, \bibinfo{person}{Yuan Cao}, {and} \bibinfo{person}{Karthik Narasimhan}.} \bibinfo{year}{2023}\natexlab{}.
\newblock \bibinfo{title}{Tree of Thoughts: Deliberate Problem Solving with Large Language Models}.
\newblock
\newblock
\showeprint[arxiv]{2305.10601}~[cs.CL]
\urldef\tempurl%
\url{https://arxiv.org/abs/2305.10601}
\showURL{%
\tempurl}


\bibitem[Yu et~al\mbox{.}(2018)]%
        {yu-etal-2018-spider}
\bibfield{author}{\bibinfo{person}{Tao Yu}, \bibinfo{person}{Rui Zhang}, \bibinfo{person}{Kai Yang}, \bibinfo{person}{Michihiro Yasunaga}, \bibinfo{person}{Dongxu Wang}, \bibinfo{person}{Zifan Li}, \bibinfo{person}{James Ma}, \bibinfo{person}{Irene Li}, \bibinfo{person}{Qingning Yao}, \bibinfo{person}{Shanelle Roman}, \bibinfo{person}{Zilin Zhang}, {and} \bibinfo{person}{Dragomir Radev}.} \bibinfo{year}{2018}\natexlab{}.
\newblock \showarticletitle{{S}pider: A Large-Scale Human-Labeled Dataset for Complex and Cross-Domain Semantic Parsing and Text-to-{SQL} Task}. In \bibinfo{booktitle}{\emph{Proceedings of the 2018 Conference on Empirical Methods in Natural Language Processing}}, \bibfield{editor}{\bibinfo{person}{Ellen Riloff}, \bibinfo{person}{David Chiang}, \bibinfo{person}{Julia Hockenmaier}, {and} \bibinfo{person}{Jun{'}ichi Tsujii}} (Eds.). \bibinfo{publisher}{Association for Computational Linguistics}, \bibinfo{address}{Brussels, Belgium}, \bibinfo{pages}{3911--3921}.
\newblock
\urldef\tempurl%
\url{https://doi.org/10.18653/v1/D18-1425}
\showDOI{\tempurl}


\end{thebibliography}

%
%

\clearpage
\section*{Appendix}

\subsection*{A. Prompt Templates}

This Appendix provides prompt templates used for each of the four predictions tasks considered in the paper. We also provide the prompt templates for the LLM-as-a-judge feedback mechanism and for the self-reflection iterations.

\vspace{1ex}

\noindent
{\large \textbf{A.1. Prediction Tasks}}

\begin{tcolorbox}[title=\textbf{Flores-200}, enhanced, breakable, colback=gray!5, colframe=gray!50!black, fonttitle=\bfseries]
Translate the following text into \{language\}. Please output only the translated text with no prefix or introduction and put in in <translation></translation> XML tags.

Text to be translated: \{source\}
\end{tcolorbox}

\begin{tcolorbox}[title=\textbf{Math500}, enhanced, breakable, colback=gray!5, colframe=gray!50!black, fonttitle=\bfseries]
What is the answer to the following math problem: \{problem\}. Make sure to always state your final answer in <answer> </answer> tags.
\end{tcolorbox}

\begin{tcolorbox}[title=\textbf{Spider}, enhanced, breakable, colback=gray!5, colframe=gray!50!black, fonttitle=\bfseries]
You are a data scientist sqlite expert. Your job is to take user questions and translate them into SQL queries. For reference, today's date is 16/04/2025. 

\{table\_name\_and\_schema\}

<instruction>
Only fetch the relevant columns for example partition is not generally required.
</instruction>

The user question is provided inside <question></question> XML tags. Aim to generate a valid sqlite query for the user question using the table above. Always provide your thinking in <reasoning></reasoning> tags and then output the SQL statement in <SQL></SQL> tags. 

Here is the question:\{question\}
\end{tcolorbox}

\begin{tcolorbox}[title=\textbf{IMDB Reviews}, enhanced, breakable, colback=gray!5, colframe=gray!50!black, fonttitle=\bfseries]
Read the following movie review. Classify the review sentiment as either positive or negative. Do not add any other words. Please output only the sentiment in <sentiment></sentiment> XML tags. Review to be classified: \{review\}
\end{tcolorbox}

\noindent
{\large \textbf{A.2. Self-Reflections and Feedback Mechanisms}}

\begin{tcolorbox}[title=\textbf{Self-Reflection}, enhanced, breakable, colback=gray!5, colframe=gray!50!black, fonttitle=\bfseries]
Please reiterate your answer by thinking step by step, making sure to state your answer at the end of the response.

\{feedback\_mechanism\_output\}

As a reminder, the original question is \{first\_user\_message\}
\end{tcolorbox}

\begin{tcolorbox}[title=\textbf{LLM-as-a-Judge Feedback}, enhanced, breakable, colback=gray!5, colframe=gray!50!black, fonttitle=\bfseries]
You are evaluating the accuracy of a response to a question. Review the following context containing both a question and answer. 

For your evaluation:
\begin{itemize}
    \item Determine if the answer is factually correct and fully addresses the question
    \item Make a binary judgment: CORRECT or INCORRECT
    \item Provide a brief justification (1-2 sentences)
    \item If you don't have enough information to make a judgment, say so
\end{itemize}

User question: \{user\_query\}

Provided response: \{context\}
\end{tcolorbox}

\subsection*{B. Extended Results}

This Appendix provides extended empirical results, including the plots depicting the impact of number of self-reflection rounds on the LLM accuracy, as well as additional Sankey diagrams revealing the transition dynamics during the self-reflection rounds on Math500.

\vspace{1ex}

\noindent
{\large \textbf{B.1. Impact of the Number of Reflections on Accuracy}}

\begin{figure}[h]\centering
\begin{subfigure}[b]{0.48\textwidth}
    \includegraphics[width=\textwidth]{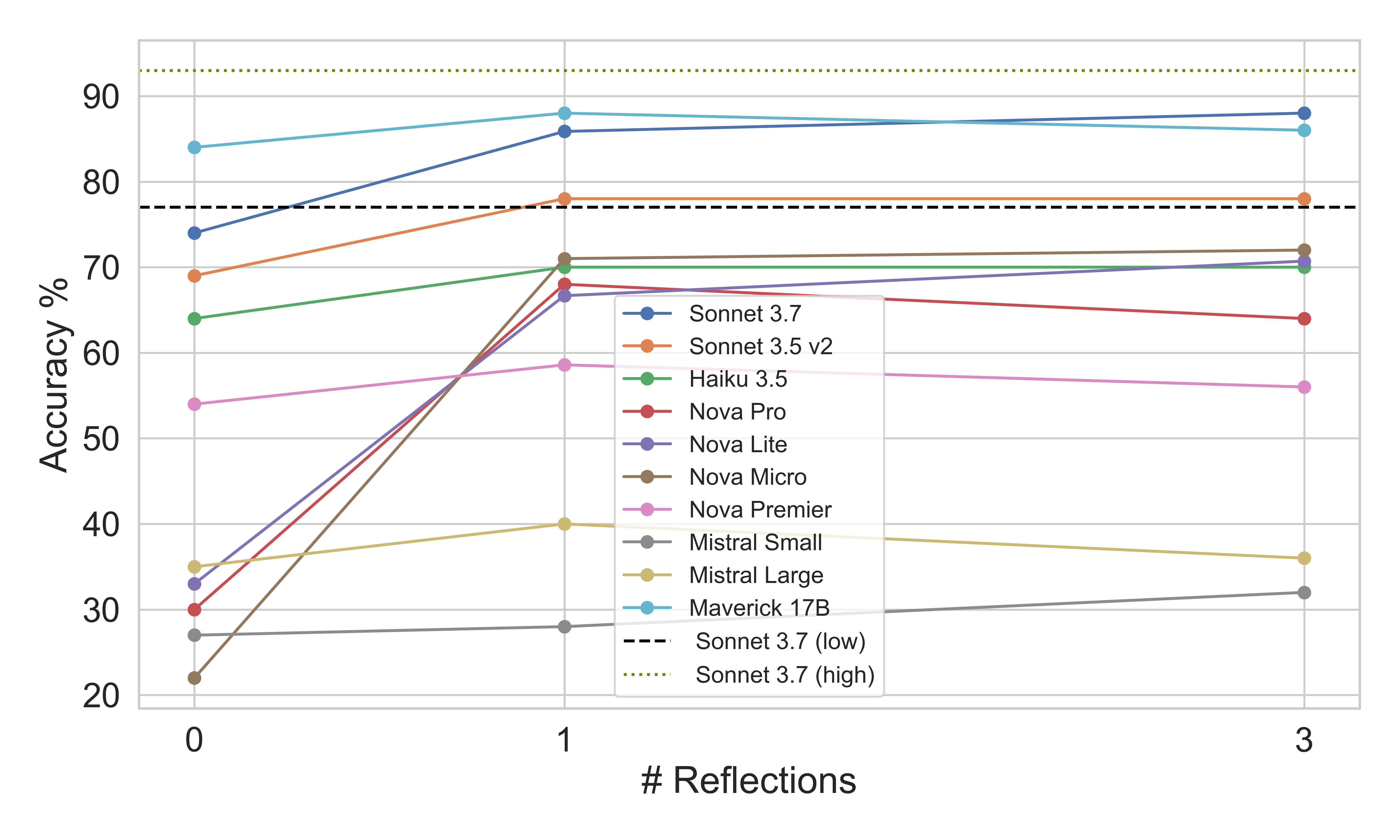}
    \caption{Math500, Reflection Impact}
\end{subfigure}
\hfill
\begin{subfigure}[b]{0.48\textwidth}
    \includegraphics[width=\textwidth]{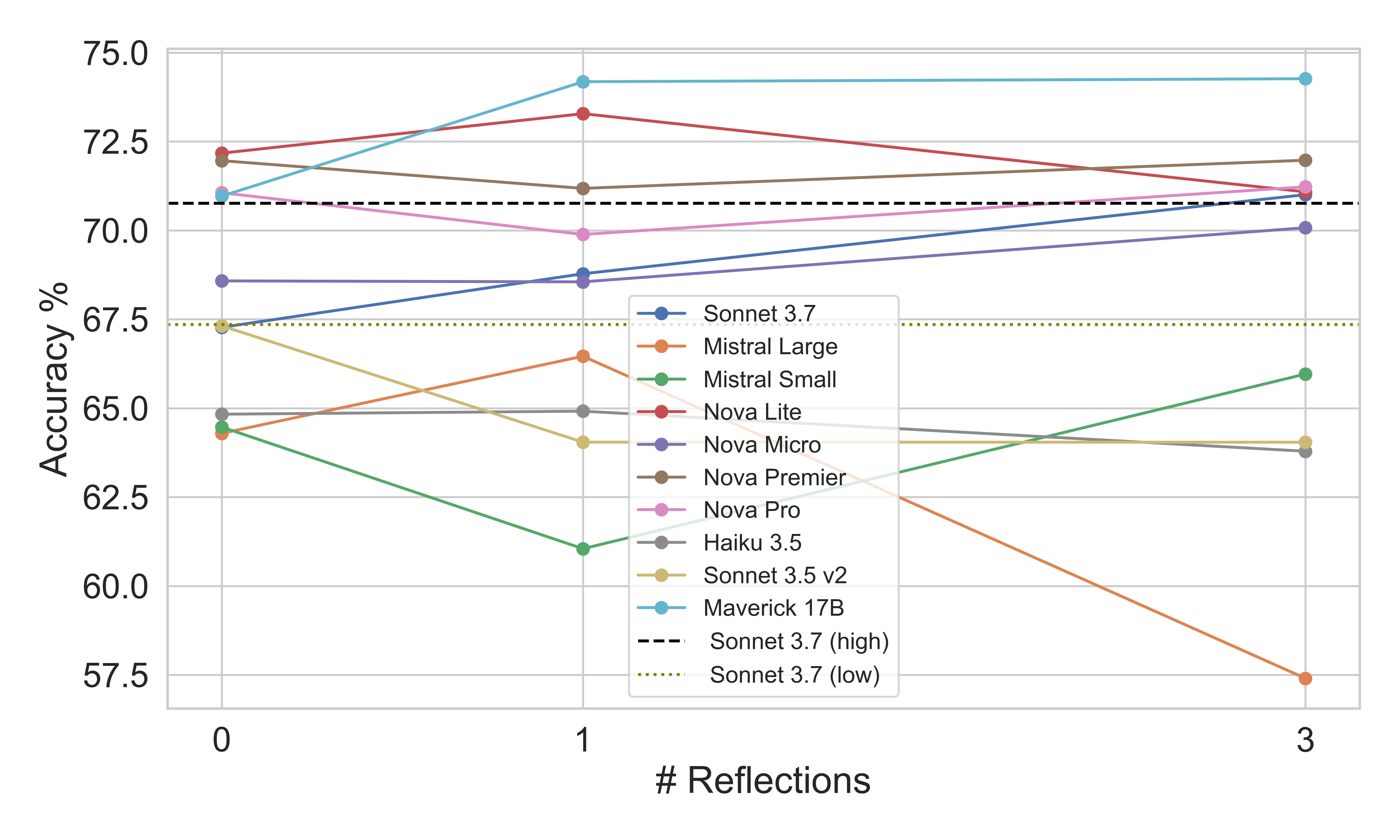}
    \caption{Text-to-SQL (Spider), Reflection Impact}
\end{subfigure}
\caption{Number of Reflections (Math500 and Spider)}
\label{fig:app_reflections_math500}
\end{figure}

\begin{figure}[h]\centering
\begin{subfigure}[b]{0.48\textwidth}
    \includegraphics[width=\textwidth]{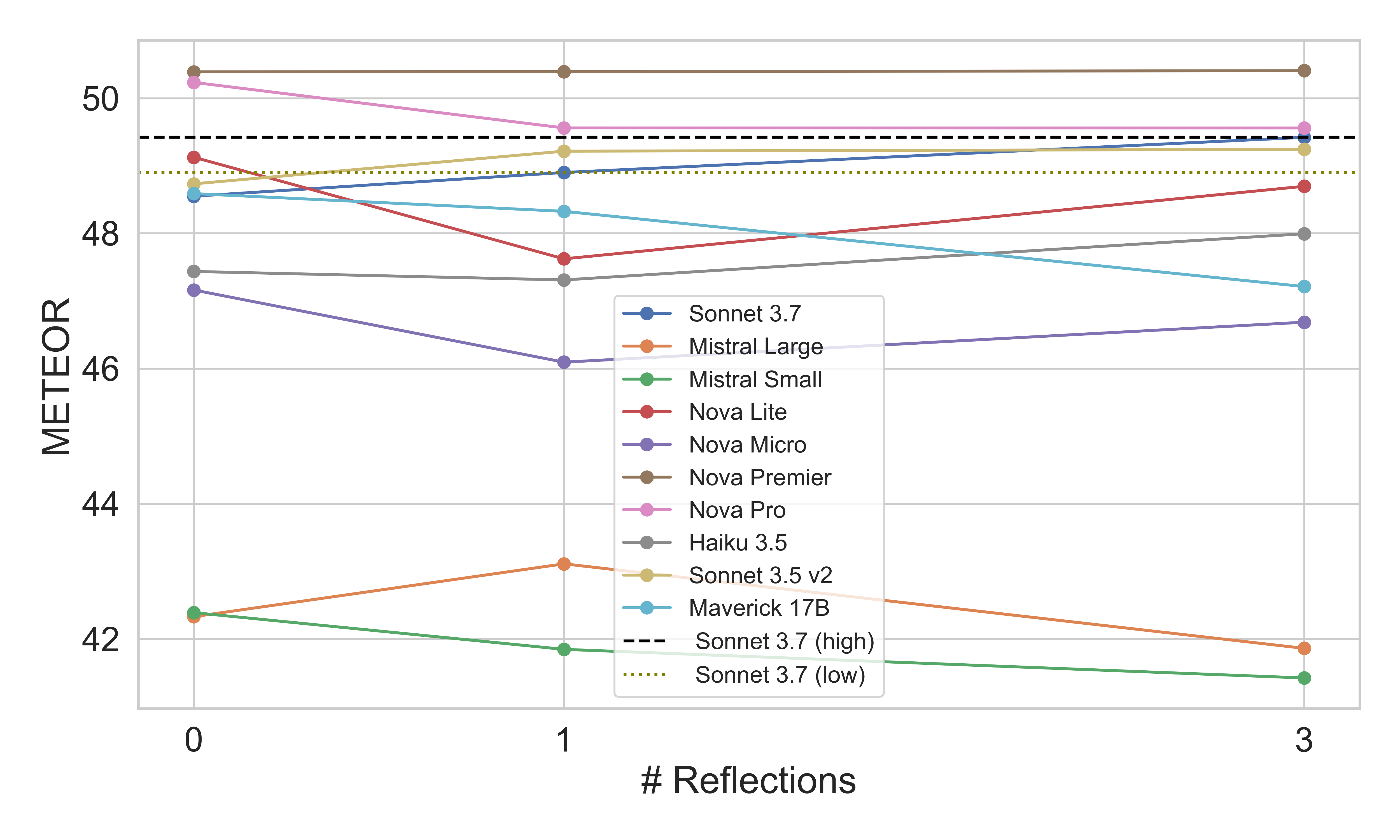}
    \caption{Translation, Reflection Impact}
\end{subfigure}
\hfill
\begin{subfigure}[b]{0.48\textwidth}
    \includegraphics[width=\textwidth]{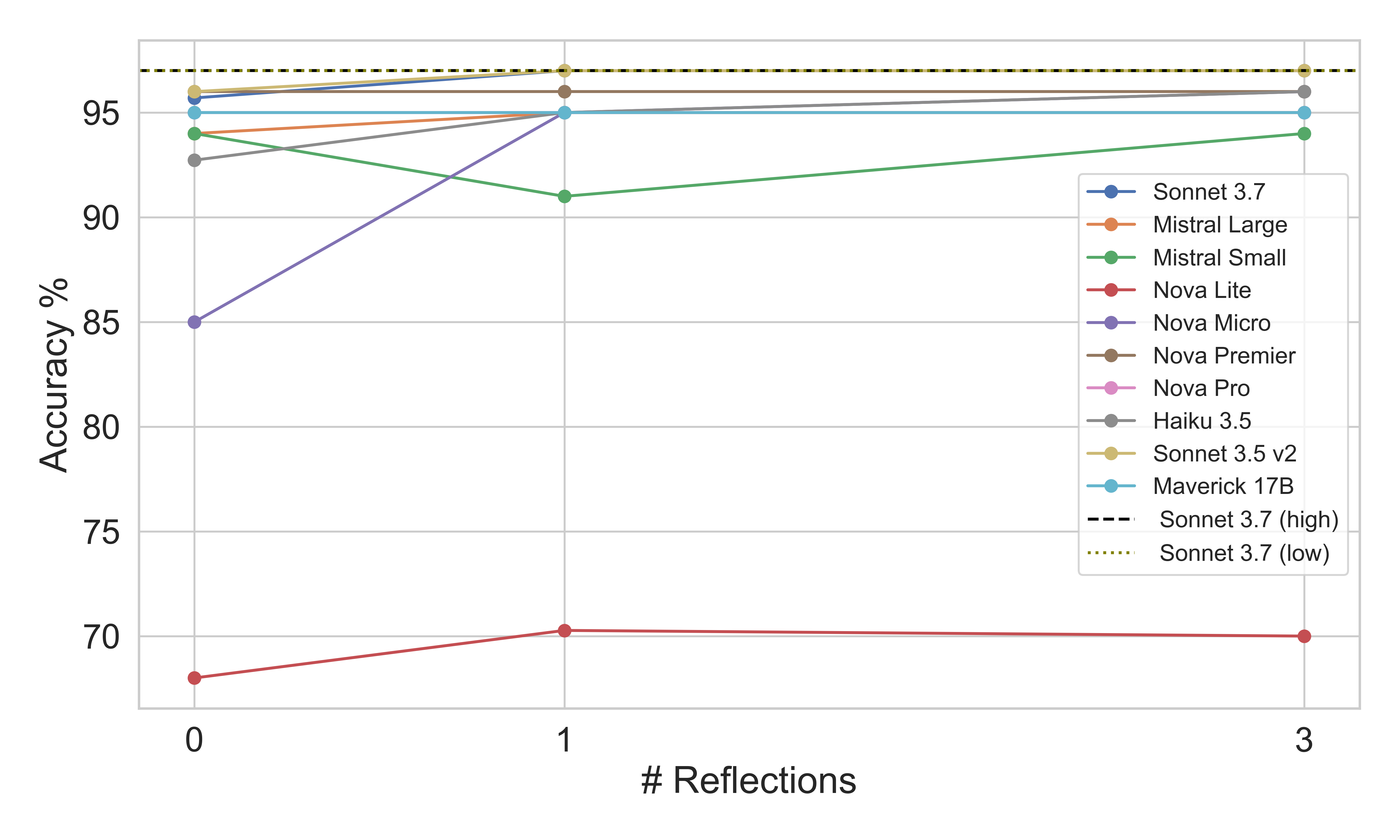}
    \caption{Sentiment Classification, Reflection Impact}
\end{subfigure}
\caption{Number of Reflections (Flores-200 and IMDB)}
\label{fig:app_transitions_imdb}
\end{figure}

\noindent
{\large \textbf{B.2. Self-Reflection Transitions}}

\begin{figure}[H]\centering
\begin{subfigure}[b]{0.49\textwidth}
    \includegraphics[width=\textwidth]{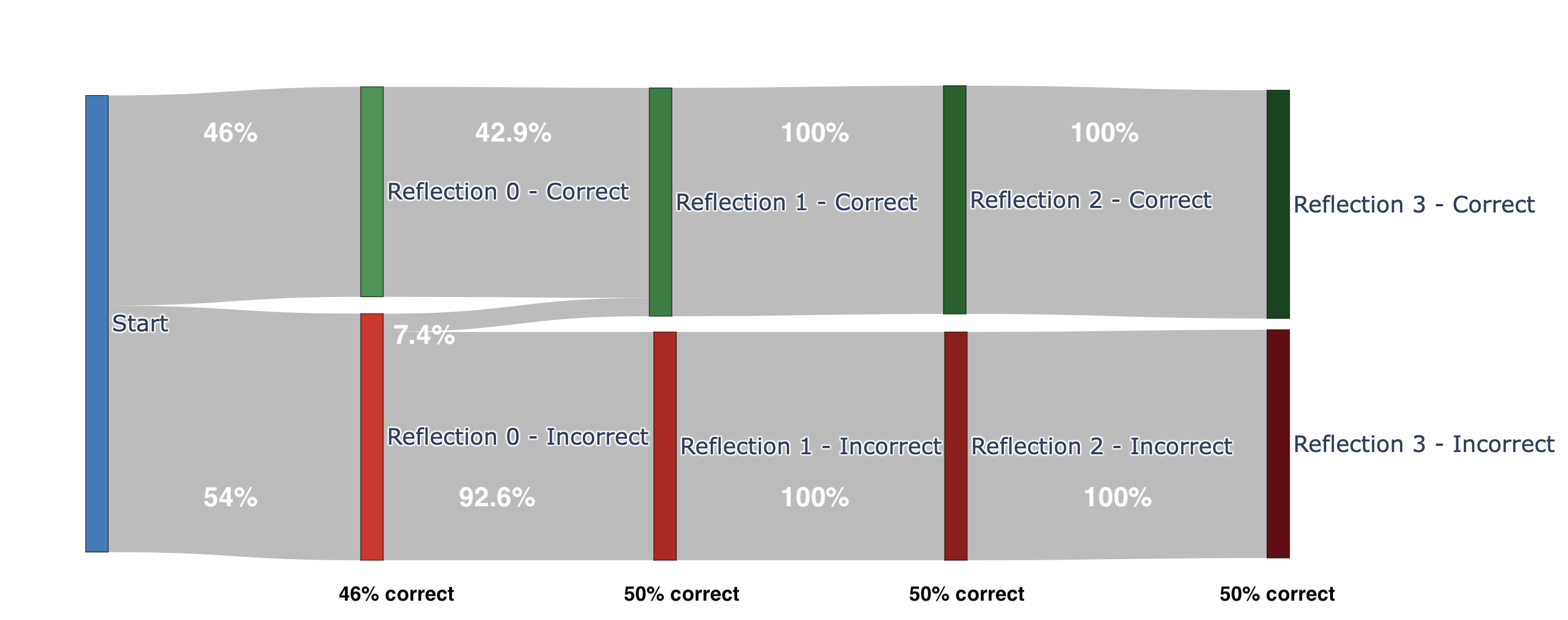}
    \caption{Math500, Amazon Nova Premier Reflection Transitions}
\end{subfigure}
\hfill
\begin{subfigure}[b]{0.49\textwidth}
    \includegraphics[width=\textwidth]{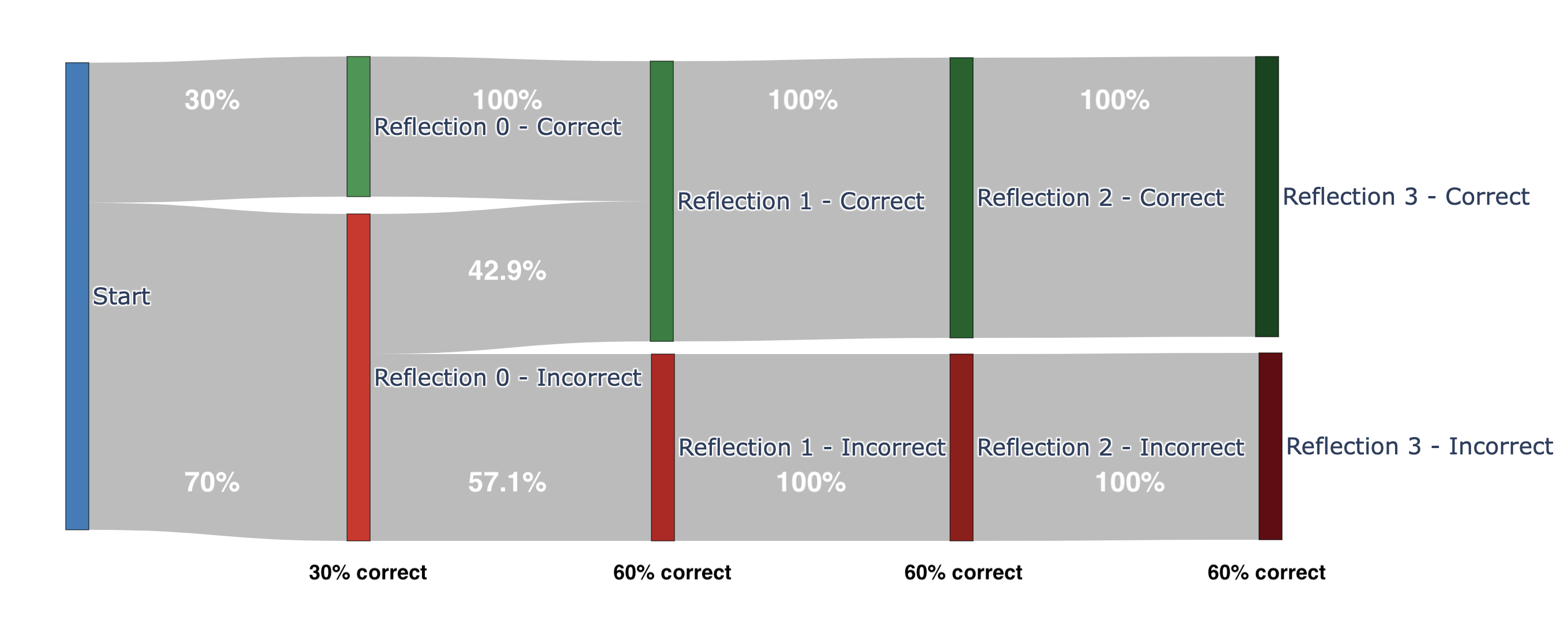}
    \caption{Math500, Amazon Nova Pro Reflection Transitions}
\end{subfigure}
\hfill
\begin{subfigure}[b]{0.49\textwidth}
    \includegraphics[width=\textwidth]{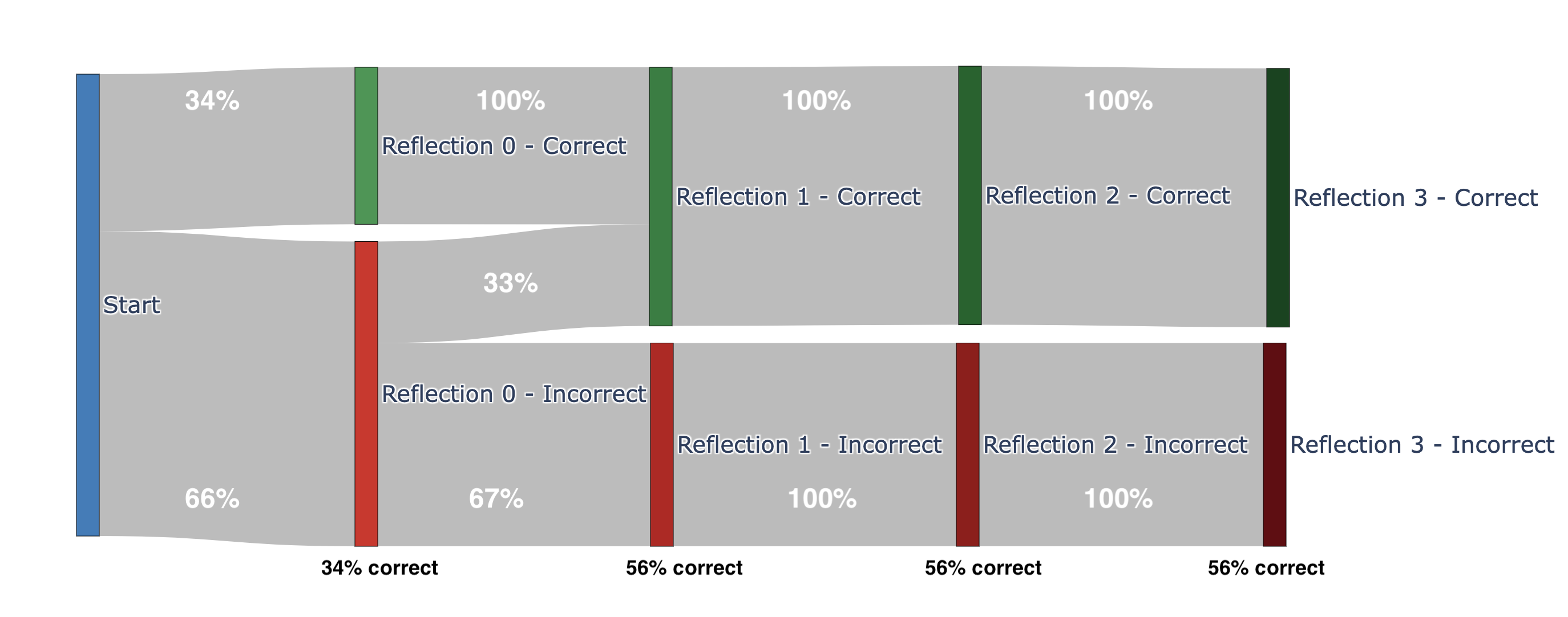}
    \caption{Math500, Amazon Nova Lite Reflection Transitions}
\end{subfigure}
\hfill
\begin{subfigure}[b]{0.49\textwidth}
    \includegraphics[width=\textwidth]{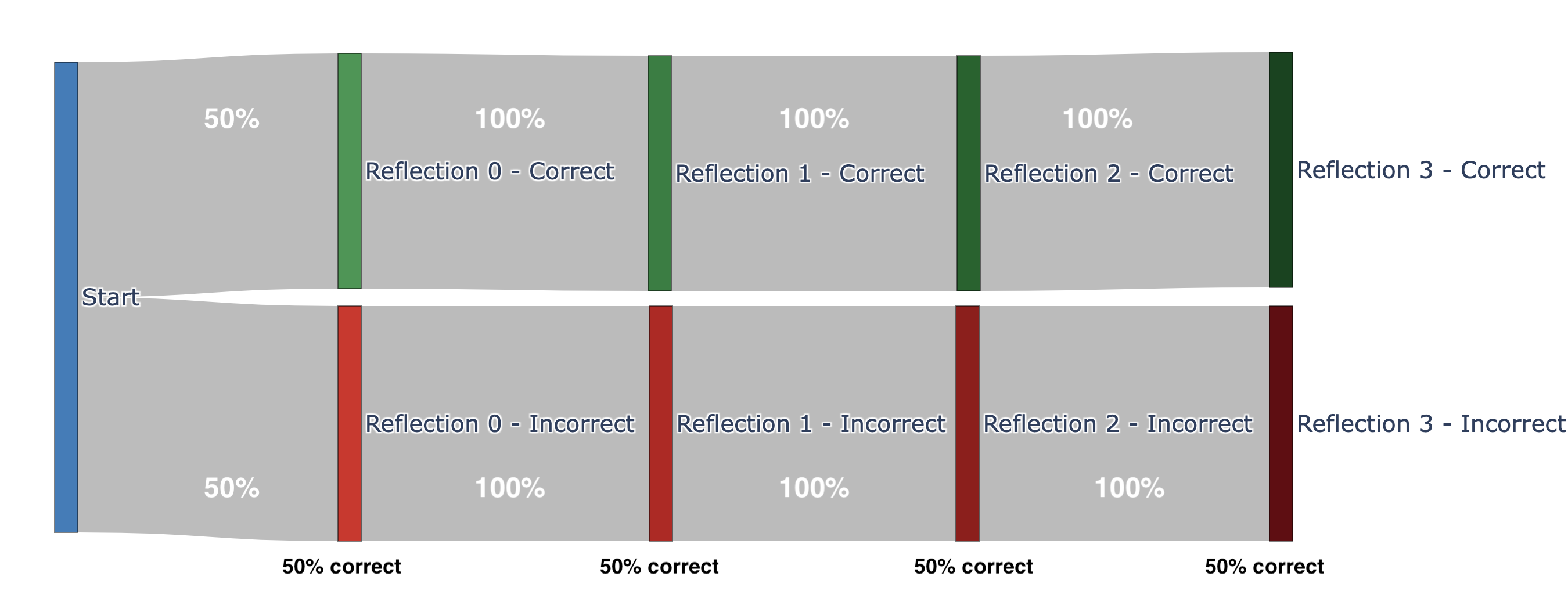}
    \caption{Math500, Anthropic Claude 3.5 Haiku Reflection Transitions}
\end{subfigure}
\hfill
\begin{subfigure}[b]{0.49\textwidth}
    \includegraphics[width=\textwidth]{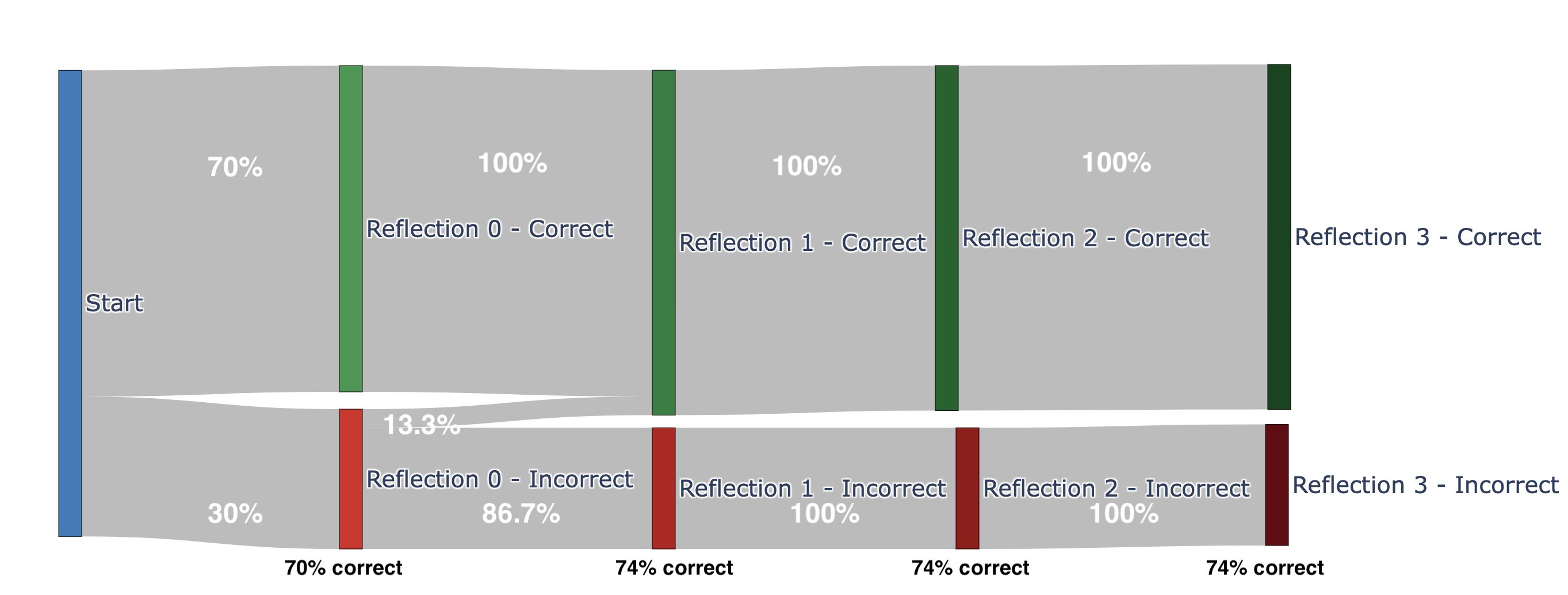}
    \caption{Math500, Anthropic Claude 3.7 Sonnet Reflection Transitions}
\end{subfigure}
\caption{Reflections Transitions (Math500)}
\label{fig:app_transitions_math500}
\end{figure}

\noindent
The Sankey diagrams (Figure \ref{fig:app_transitions_math500} a-e) provide detailed visualisation of reflection pathways for additional models beyond Claude Sonnet 3.5 v2 and Amazon Nova Micro discussed in the main text. These diagrams reveal consistent patterns across model families while highlighting unique characteristics. Models with varying initial accuracy (34\%-70\%) all demonstrate perfect retention of correct answers through subsequent reflection stages; a pattern consistent across all tested LLMs. For models with moderate initial performance (46\%-50\%), we observe that the first reflection stage provides the most substantial correction opportunity, with 42.9\%-67\% of initially incorrect responses remaining incorrect after Reflection 0, while subsequent reflections yield minimal improvements. This mirrors Amazon Nova Micro's behavior described in the main text. In contrast, models with higher initial accuracy (70\%) show more nuanced improvement patterns, with 13.3\% of initially incorrect responses being corrected at Reflection 0 and accuracy stabilising at 74\%—similar to Claude Sonnet 3.5's incremental improvement pattern. These findings reinforce our main conclusion that smaller models primarily benefit from initial self-correction, while more capable models can leverage both strong foundational performance and iterative improvement through extended reflection processes.

\newpage
\noindent

\begin{figure*}[ht]\centering
\includegraphics[width=0.78\textwidth]{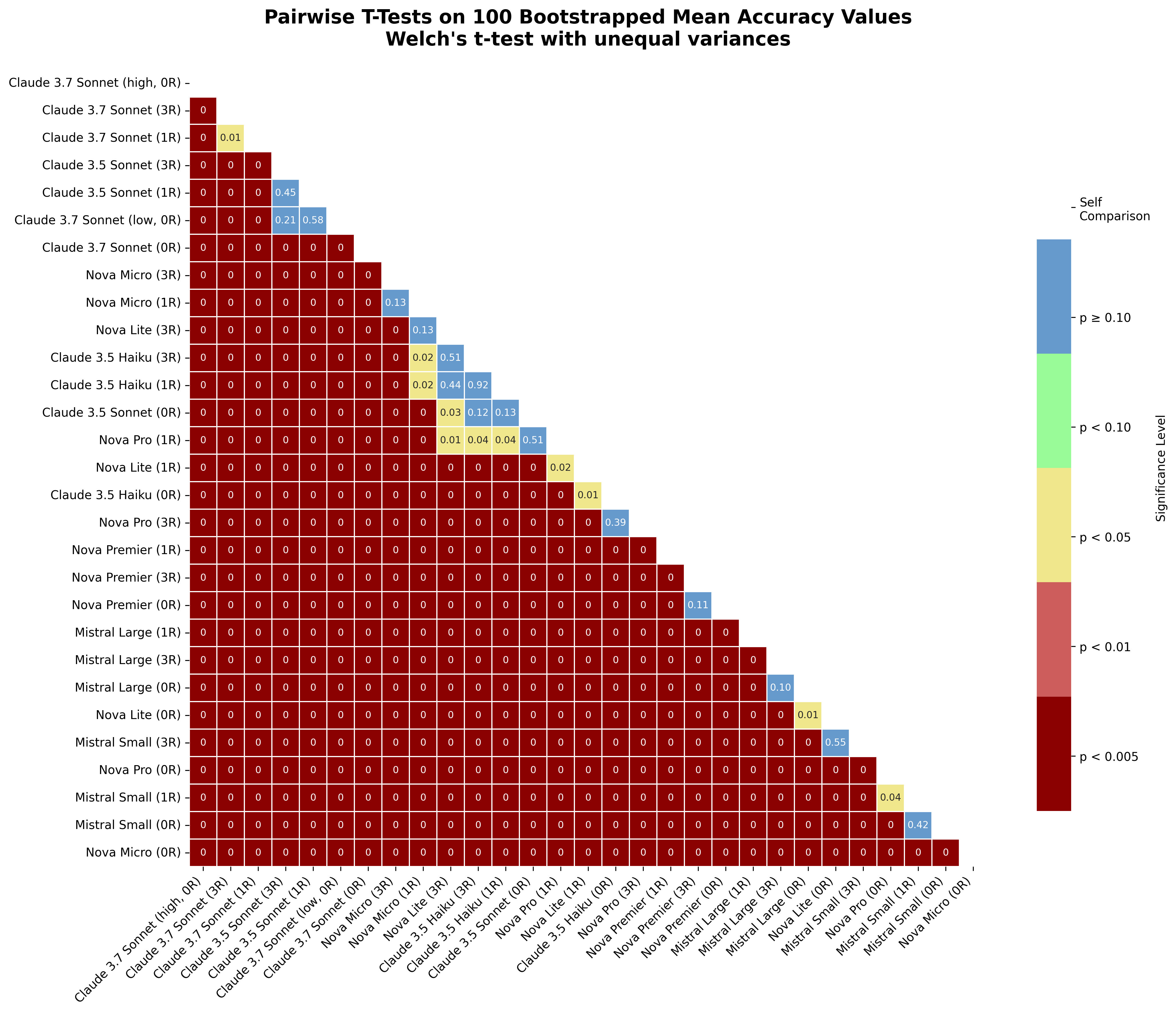}
\caption{Math500, Pairwise T-Test P-Values}
\label{fig:ttests}
\end{figure*}

\newpage

\noindent
{\large \textbf{B.3. Statistical Significance Tests}}

\noindent
To validate the statistical significance of the accuracy differences observed in Section 4, we draw 100 bootstrap samples of the individual LLM generations for each model and inference-time configuration. Next, we calculate accuracy scores per configuration on each of the bootstrap samples, which yields a distribution of 100 accuracy scores for each configuration. Finally, we run pairwise statistical tests comparing the accuracy distributions.

We use pairwise Welch's t-tests with unequal variances to compare mean accuracy values of different configurations. Figure \ref{fig:ttests} shows test p-values across all model configurations on Math500, sorted by the average accuracy. Out of 496 configuration pairs, only 26 accuracy differences are not significant at 1\% level, including different variations of Claude 3.5 Haiku compared to each other and Nova Pro with 1 self-reflection. 

Beyond the t-tests, significance of the accuracy differences is also confirmed by the Friedman test, which indicates there are  significant differences between at least some of the 32 model configurations and rejects the null hypothesis that all models perform equally. Nemenyi post-hoc tests indicate that 71\% of the pairwise accuracy differences are statistically significant, including the models on the efficiency frontiers.

\vspace{1ex}

\noindent
{\large \textbf{B.4. Prompt Caching}}

\noindent
Prompt caching, as described generally in \cite{gim2024promptcachemodularattention}, is a set of techniques for caching computed model states so they can be re-used over future invocations of an LLM. Amazon Bedrock has released a prompt caching feature which allows users to set cached checkpoints during their conversation history and then save on the cost of recomputing these past messages. This capability is often used to cache very long initial system prompts or initial context which is used across multiple interactions with the LLM. Additionally, our results such as Figure \ref{fig:imdb} have shown that while self-reflection can bring improved performance, the additional cost and latency can hurt the feasibility of integrating these techniques.

Self-reflection, as we have defined in the earlier sections, has the potential to benefit from the prompt caching approach as we are frequently asking the model to reflect on past messages and revise the response. This is different to reasoning models such as Claude Sonnet 3.7, as their thought process is typically contained within the internal thinking tokens and not explicitly defined as sequences of messages in a conversation chain, preventing the use of prompt caching features available in Amazon Bedrock. 

\begin{figure}[h]\centering
\begin{subfigure}[b]{0.49\textwidth}
    \includegraphics[width=\textwidth]{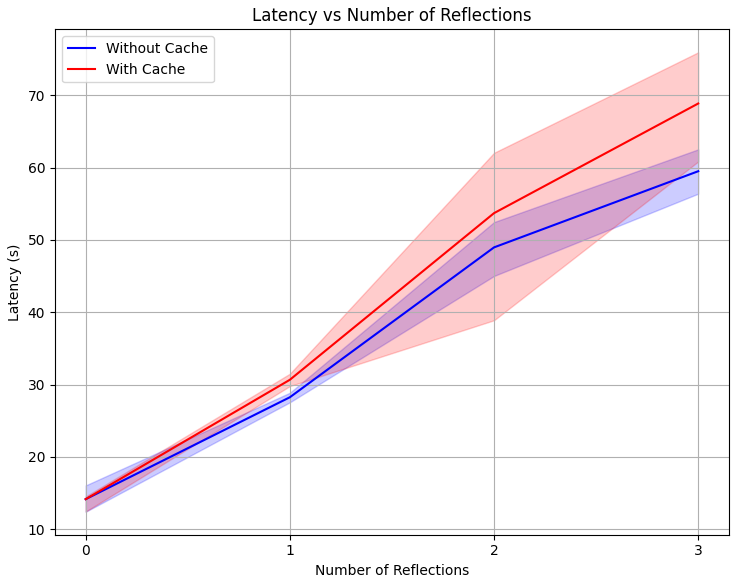}
    \caption{Latency Trade-off for Prompt Caching and Self-reflection}
    \label{fig:cache_latency}
\end{subfigure}
\hfill
\begin{subfigure}[b]{0.49\textwidth}
    \centering
    \includegraphics[width=\textwidth]{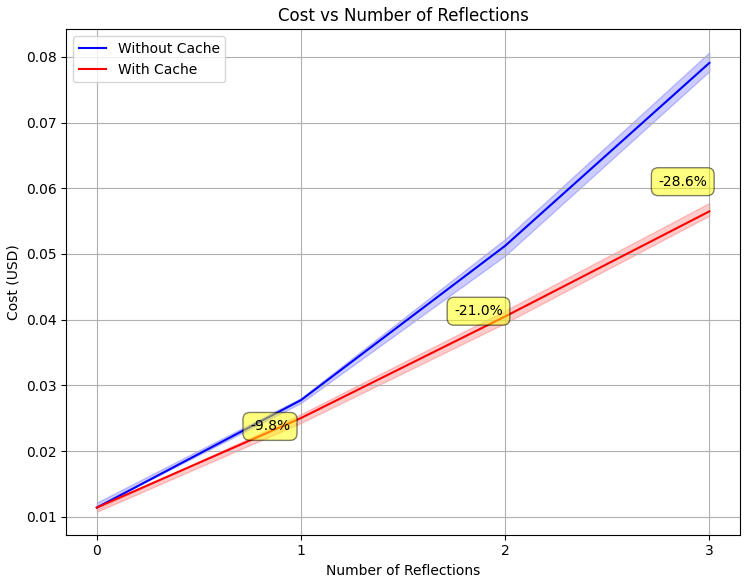}
    \caption{Cost Trade-off for Prompt Caching and Self-reflection. The percentage difference is the difference in mean cost.}
    \label{fig:cache_cost}
\end{subfigure}
\caption{Prompt Caching cost (\$) and latency trade-off results for a sampled Text-to-SQL prompt, repeated over 3 distinct rounds of generation with the mean and variance shown as $\mu \pm \sigma$}
\label{fig:prompt_caching}
\end{figure}

To analyse this trade-off further, we explore the differences in cost and latency across multiple rounds of self-reflection with and without leveraging the prompt caching feature in Amazon Bedrock. Figure \ref{fig:prompt_caching} shows the cost and latency for a typical sequence of self-reflection rounds, with the model prompted to solve a Text-to-SQL question using an initial prompt size of approximately 1,000 tokens. Interestingly, Figure \ref{fig:cache_latency} shows that prompt caching combined with self-reflection has minimal benefits in terms of reducing the latency. We hypothesise that this could be due to the additional overhead of reading from cache databases being approximately equal to the latency required to generate the relatively minimal 100's of tokens. However, Figure \ref{fig:cache_cost} demonstrates that integrating self-reflection with prompt caching can being up to 28\% reduction in cost when sampling over 3 rounds of reflection. 

This method allows for more cost effective, linear scaling of self-reflection where only the incremental cost of additional output tokens is expensed with each round of reflection. For practitioners, it implies that leveraging self-reflection techniques can be more valuable with the LLMs and model providers that support prompt caching, as it offsets a significant part of additional costs on reflection rounds. We see potential for these techniques to grow in impact as more model providers enable improved prompt caching mechanisms in the future. 

\end{document}